\documentclass[runningheads]{llncs}
\usepackage{graphicx,pbox}
\usepackage{rotating,amsmath}
\usepackage{algorithm}
\usepackage{algcompatible}
\usepackage{booktabs}
\usepackage{multirow,capt-of}
\usepackage{tikz}
\usepackage{comment}
\usepackage{amsmath,amssymb} 
\usepackage{color}

\usepackage[accsupp]{axessibility}  

\begin{document}
\pagestyle{headings}
\mainmatter
\def\ECCVSubNumber{2764}  

\title{No Token Left Behind: Explainability-Aided Image Classification and Generation} 

\titlerunning{No Token Left Behind}
%
\author{Roni Paiss\inst{1} \and
Hila Chefer\inst{} \and
Lior Wolf\inst{}}
\authorrunning{R. Paiss et al.}
%

\institute{The Blavatnik School of Computer Science, Tel Aviv University}
\maketitle
\footnotetext[1]{Work was done while the author was also working at Apple.}

\begin{abstract}
The application of zero-shot learning in computer vision has been revolutionized by the use of image-text matching models. The most notable example, CLIP, has been widely used for both zero-shot classification and guiding generative models with a text prompt. However, the zero-shot use of CLIP is unstable with respect to the phrasing of the input text, making it necessary to carefully engineer the prompts used. 
We find that this instability stems from a selective similarity score, which is based only on a subset of the semantically meaningful input tokens.
To mitigate it, we present a novel explainability-based approach, which adds a loss term to ensure that CLIP focuses on all relevant semantic parts of the input, in addition to employing the CLIP similarity loss used in previous works. 
When applied to one-shot classification through prompt engineering, our method yields an improvement in the recognition rate, without additional training or fine-tuning. Additionally, we show that CLIP guidance of generative models using our method significantly improves the generated images.
Finally, we demonstrate a novel use of CLIP guidance for text-based image generation with spatial conditioning on object location, by requiring the image explainability heatmap for each object to be confined to a pre-determined bounding box. Our code is available at \url{https://github.com/apple/ml-no-token-left-behind}.
\end{abstract}

\section{Introduction}
\label{sec:intro}
State-of-the-art computer vision models are often trained as task-specific models that infer a fixed number of labels. In contrast, \cite{radford2021learning} have demonstrated that by training an image-text matching model that employs Transformers for encoding each modality, tens of downstream tasks can be performed without further training (``zero-shot''), with {comparable accuracy} to the state of the art~\cite{radford2021learning}. 
Due to its zero shot capabilities and its semantic latent space, CLIP~\cite{radford2021learning} has been widely used in recent research to guide pretrained generative networks to create images according to a text prompt~\cite{gal2021stylegannada,vqgan+clip,Liu2021FuseDreamTT} and edit an input image according to a text description~\cite{kim2021diffusionclip,patashnik2021styleclip,Avrahami2021blendeddiffusion}.

While CLIP shows great promise in zero-shot image classification tasks and generative network guidance, it suffers from instabilities that often lead to biased similarity scores between non-matching text and image pairs. To mitigate these instabilities, \cite{radford2021learning}  suggest a prompt engineering technique that averages the embeddings of multiple textual templates. 
Since CLIP summarizes the relation between a given image and a given text with a single similarity score, it can present a myopic behavior and focus on specific elements within the sentence and/or the image. 
In order to alleviate this issue, it is necessary to rely on an additional signal. In this work, we propose using the explainability maps 
to steer the optimization process towards solutions that rely on the relevant parts of the input, and away from solutions that focus on irrelevant parts, or on a small subset of the relevant parts.
We explore two domains in which we guide CLIP to account for the important tokens in an input prompt: one-shot classification and zero-shot text-based image generation. For one-shot classification, we incorporate a loss based on the explainability scores of the class name in the prompt engineering method proposed by~\cite{zhou2021coop}. Our results demonstrate that guiding the prompt engineering process using explainability improves performance in both the class-agnostic and the class-specific cases. 
In the domain of image editing guided by text, we employ a similar explainability-based loss. This loss allows the generative network to avoid local minima caused by focusing on irrelevant words in the input text, by requiring the explainability scores of important tokens to be high. We demonstrate that our method significantly improves the generated images.
Additionally, by applying similar principles to the image-side relevancy map, we use the obtained heatmaps to facilitate CLIP-guided text-based image generation with spatial conditioning. As far as we can ascertain, we are the first to present a spatial layout to image method using CLIP. As we demonstrate, a straightforward application of the similarity score over {a requested} bounding box does not guarantee that the entire object will be contained within that bounding box. When relying on the explainability heatmap, our method helps ensure that the object does not deviate from the provided bounding box.

\section{Related Work}
\textbf{Zero-shot classification\quad} Zero-shot classification in computer vision usually refers to a model's ability to generalize to unseen labels. While several works used weakly labeled Google images as training data~\cite{Berg_animals,Chen2015WeblySL,Fergus,Rubinstein,Tang,Vijayanarasimhan,Wang} the method of \cite{Li_2017_ICCV} was perhaps the first to study zero-shot transfer learning to unseen datasets, which is a broader approach to zero-shot classification. This approach was adopted by CLIP~\cite{radford2021learning}, which trains an image-text matching engine using an image encoder and a text encoder, via contrastive learning. The image encoder architectures used are either ViT~\cite{dosovitskiy2020image} or ResNet~\cite{he2016deep}, and the text encoder is based on a Transformer~\cite{vaswani2017attention} with the modifications of \cite{Radford2019LanguageMA}.  CLIP was trained on a set of $400M$ matching image-text pairs, and showed remarkable zero-shot capabilities on the ImageNet dataset~\cite{Deng2009ImageNetAL}.  Following CLIP, \cite{zhou2021coop} proposed few-shot prompt engineering to enhance CLIP's classification accuracy on unseen datasets. Their approach opts to learn the textual templates fed to CLIP rather than manually engineering them, as originally done by \cite{radford2021learning}.
As we show, CLIP-guided optimization methods such as CoOp~\cite{zhou2021coop} tend to focus on a sparse set of tokens in the input, and often neglect important parts of it. Our work attempts to mitigate this issue by applying an explainability-based loss.

\smallskip
\noindent\textbf{CLIP-guided Generation\quad} Following the success of CLIP in zero-shot classification, several works have used the similarity scores produced by CLIP to guide pretrained generative networks. These methods usually construct a similarity-based loss, encouraging a generator to produce the output with the desired semantic attributes. 
Some of the applications of CLIP for guiding generative networks include image manipulation~\cite{patashnik2021styleclip}, image essence transfer~\cite{chefer2021imagebased}, style transfer~\cite{gal2021stylegannada,Zhu2021MindTG}, 3D object style editing~\cite{Michel2021Text2MeshTN}, and image captioning~\cite{Tewel}.
While demonstrating great capabilities, we show that methods such as StyleCLIP and VQGAN+CLIP are limited by the tendency of CLIP to sometimes ignore meaningful parts of the text. Our explainability-based loss addresses this issue.

\smallskip
\noindent\textbf{Transformer Explainability\quad} Many methods were
suggested for generating a heatmap that indicates local relevancy, given an input image and a CNN~\cite{selvaraju2017grad,binder2016layer,lundberg2017unified,shrikumar2016not}. However, the literature on Transformer explainability is relatively sparse and most methods focus on pure self-attention architectures~\cite{abnar2020quantifying,Chefer_2021_CVPR}. The recent method of \cite{Chefer_2021_ICCV}, which we employ in this work, is the first to also offer a comprehensive method for bi-modal networks. 

\section{Method}
\label{sec:method}
We begin by describing how to produce relevance values for each word and image patch using CLIP. We then describe how our method is applied to one-shot classification via prompt engineering and to zero-shot image generation.

\subsection{Explainability}
\label{sec:explainability}
Given a pair of text $t$ and image $i$, CLIP produces a score $\text{CLIP}(t,i)$, which determines how semantically similar the textual description $t$ is to the image $i$. The goal of the explainability method is to produce a relevance score for each input text token and image patch in the computation of the similarity score $\text{CLIP}(t,i)$. The score of each token should reflect its impact on the similarity score, and input tokens with the greatest influence on the similarity score should receive a high relevancy score and vice versa.
We employ the method described in \cite{Chefer_2021_ICCV} (details in App.~\ref{app:method}) to produce a relevance score for each image patch and text token, given the calculated similarity score.
As the relevance scores are calculated per token, we define the relevance score of a word to be the maximal relevance score among its tokens. For each word $W=w_1,...,w_n$, where $w_1,...w_n$ are the tokens it contains, we define its relevance score $\mathcal{R}_{expl}(W)$ as $\mathcal{R}_{expl}(W) = \max_{k\in w_1,...,w_n}\mathbf{R}[k]$,
where $\mathbf{R}[k]$ is the relevance score of \cite{Chefer_2021_ICCV}.

\subsection{Prompt engineering}
\label{sec:prompt}
Image classification is the task of assigning a label from a set of possible classes $c\in C$ to an input image $i$.
In order to adapt CLIP to different classification tasks, \cite{radford2021learning} propose employing prompt templates with each possible class $c\in C$ inserted, e.g. ``A photo of a \{label\}.'' These templates are necessary because in the process of CLIP's pre-training most textual inputs are full sentences rather than single words.
Let $i$ be the input image to be classified, and let $t$ denote the textual template. $t(c)$ denotes the template $t$, where the \{label\} placeholder was replaced by the class name $c$. CLIP scores per class are obtained using the similarity score between the input image and the class-template as follows:
\begin{align}
\label{eq:softmaxclip}
      \Pr(\text{output}=c | i)=\frac{e^{\text{CLIP}(t(c), i)}}{\sum_{c'\in C}e^{\text{CLIP}(t(c'), i)}}.
\end{align}

\cite{zhou2021coop} replace the manual selection of the textual templates with a few-shot learning of it. Given the desired template length $M$, the template $t(\text{label})=v_1,..., v_i,\{\text{label}\} , v_{i+1},..., v_M$ is optimized with a cross-entropy loss using Eq.~\ref{eq:softmaxclip} to extract the distribution.
Note that the learned templates are prone to overfitting, due to the small number of example images for each label, which can result 
in prompts that describe distinctive parts of the images that are irrelevant to their class, yielding a similarity score $\text{CLIP}(t(c), i)$ that is not based on the class name. This problem is most prominent in the one-shot scenario where the prompts are optimized based on a single image per class.
To help avoid this phenomenon, our method employs a novel explainability-based loss.  
For each class $c\in C$ and image $i$, the similarity score $\text{CLIP}(t(c), i)$ is produced, and then a normalized explainability score is calculated. This score reflects the relevance of the class $c\in C$ to the similarity of the template and the image:
\begin{align}
      \mathcal{S}_{expl}(c) = \frac{\max_{W\in c} 
\mathcal{R}_{expl}(W)}{\sum_{U\in t(c)/c}\mathcal{R}_{expl}(U)}
\end{align}
where, as above, $W\in c$ are the words comprising the label $c\in C$.
The score $\mathcal{S}_{expl}(c)$ encapsulates the impact that the class name $c$ has on the calculated similarity score $\text{CLIP}(t(c), i)$, in comparison to all other words in the sentence.
Our explainability-based loss is, therefore:
\begin{align*}
      \mathcal{L}_{expl} = \lambda_{expl}\left(  -\mathcal{S}_{expl}(c_{gt}) + \sum_{c\neq c_{gt}\in C} \mathcal{S}_{expl}(c) \right)
\end{align*}
where $c_{gt}$ is the ground truth label, and $\lambda_{expl}$ is a hyperparameter. The first term, $-\mathcal{S}_{expl}(c_{gt})$ encourages the similarity score for the ground truth class to be based on the class name tokens, in order to avoid focusing on other, irrelevant tokens. The second term, $ \sum_{c\neq c_{gt}\in C} \mathcal{S}_{expl}(c)$ encourages the similarity score for all the counterfactual classes to be based on tokens that are not the false class name, since the class name does not correspond to the image.

\subsection{Zero-shot text-guided image manipulation}
\label{sec:manipulation}
Recent research~\cite{patashnik2021styleclip,kim2021diffusionclip} demonstrates that CLIP can be effective for guiding generative networks that synthesize and edit images, by maximizing the similarity score between the desired text and image. As pointed out by~\cite{Liu2021FuseDreamTT}, methods that integrate a pre-trained generator with the CLIP score to allow text-based editing suffer from instabilities, since similarity-based optimization often reaches local minima that do not reflect the desired semantic meaning of the input query. 

\begin{figure}[t]
  \centering
  \begin{tabular}{cccc@{~~~}l}
   & Input & StyleCLIP & Ours& Colormap\\

\centering{\begin{footnotesize}\pbox[b][][b]{2cm}{Input or\\Output\\~\\~}\end{footnotesize}}
&
\includegraphics[width=0.17\linewidth, clip]
{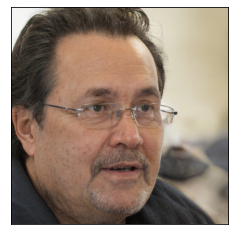}&
\includegraphics[width=0.17\linewidth,clip] 
{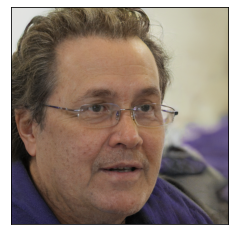}&
\includegraphics[width=0.17\linewidth,clip] 
{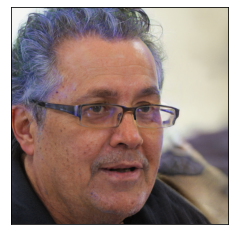}&\\

\centering{\begin{footnotesize}\pbox[b][][b]{2cm}{~~Image\\heatmap\\ ~ \\} \end{footnotesize}}&
\includegraphics[width=0.16\linewidth,clip] 
{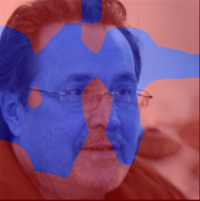}&
\includegraphics[width=0.16\linewidth,clip] 
{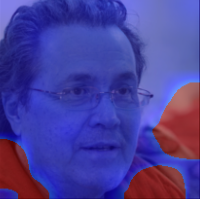}&
\includegraphics[width=0.16\linewidth,clip] 
{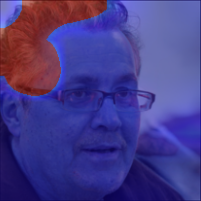} & 
\begin{footnotesize}\pbox[b][][b]{2cm}{red=high relevancy\\blue=low relevancy\\}\end{footnotesize}\\
\\
\centering{\begin{footnotesize}\pbox{20cm}{~~Text\\heatmap} \end{footnotesize}}&
\includegraphics[width=0.17\linewidth,clip] 
{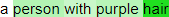}&
\includegraphics[width=0.17\linewidth,clip] 
{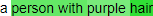}&
\includegraphics[width=0.17\linewidth,clip] 
{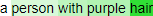} & \centering{\begin{footnotesize}\pbox{20cm}{green=high \\ relevancy} \end{footnotesize}}
\end{tabular}
\caption{Manipulations for ``A person with purple hair''. StyleClip~\cite{patashnik2021styleclip} produces a manipulation that is not consistent with the semantic meaning of the prompt, and the color of the person's shirt and the background are altered. Our method generates an output that is faithful to the input text query, and the high values of the explainability heatmaps are much more correlated with the prompt.
\label{fig:motivation}}
\end{figure}

As shown in Fig.~\ref{fig:motivation}, this shortcoming is often manifested in the explainability scores, and is caused by similarity scores relying only on a partial subset of the semantic tokens in the text query. Thus, our method leverages the explainability scores, to ensure that the output similarity scores are derived from all of the tokens that determine the semantic meaning of the input text.
Given a pre-trained generator $G$ (a mapping from a latent vector $z$ to an output image), an input image $i$, and an input text $t$, the goal of the optimization algorithm $A$ is to find a vector $A(G, i,t) = z$ such that $G(z)$ combines the visual appearance of image $i$ with the description in the text query $t$. 
In order to assess the correlation between the manipulation $G(z)$ and the desired semantic property in $t$, the algorithm $A$ uses the CLIP similarity score between the manipulated image $G(z)$ and the textual prompt $t$ as a loss term $     \mathcal{L}_{similarity} = -\text{CLIP}(G(z), t)$. 
This loss is applied in addition to other loss terms that lead to a high visual similarity between $i$ and $G(z)$.

As mentioned, $\mathcal{L}_{similarity}$ can produce biased similarity scores, which do not reflect the semantic meaning in $t$, due to focusing only on a subset of the semantically important words in $t$. Our method remedies this bias by adding a loss term that encourages CLIP to attend to all semantic words in $t$.

Let $S$ be the set of semantic words in $t$. 
Since the textual prompts for image editing are of the format: "a person/ man/ woman with \{\textit{description}\}" or of the format "a \{\textit{description}\} person/ man/ woman", the set $S$ is considered to be the words that comprise the description. Our method adds the following explainability-based loss term to the optimization algorithm $A$:
\begin{align}
      \label{eq:expl_gen_loss}
      \mathcal{L}_{expl} = -\lambda_{expl}\frac{1}{|S|}\left(\sum_{s \in S} \mathcal{R}_{expl}(s) \right),
\end{align}
where $\lambda_{expl}$ is a hyperparameter. For example, in Fig.~\ref{fig:motivation}, the set of semantic words is defined to be: $S=\{``purple'', ``hair''\}$. This helps the optimization process to favor results where the similarity score is based on the hair color of the subject of the image. As can be seen in the figure, when our loss is not applied, the optimization results in coloring the shirt and the background.

\smallskip\noindent\textbf{Choosing $\lambda_{expl}$\quad} Our modified optimization algorithm has an additional hyperparameter  $\lambda_{expl}$. Since CLIP-based optimization is sensitive to the choice of hyperparameters, it is better to set them based specifically on the input image $i$, input text $t$, and generator $G$. 
In order to provide an automatic mechanism for choosing $\lambda_{expl}$, we consider a range of possible $\lambda_{expl}$, and choose the value of $\lambda_{expl}$ for which the similarity predicted by CLIP for the generated image and the input text is maximal. Note that after applying our method, the similarity scores become more stable, as they consider all semantic tokens in the input.

\subsection{Zero-shot text-to-image with spatial conditioning}
\label{sec:l2i}
While the textual descriptions provided to CLIP can include the spatial positioning of objects, images generated by optimizing CLIP similarity score with such texts tend not to follow these spatial restrictions, as shown in Fig~\ref{fig:l2i}. We attribute this to the nature of the task CLIP was trained on, which is predicting how similar a given image is to a given text. 
The existence of matching entities in both inputs is a stronger indication of their similarity than the positions of the entities. 
This intuition is reflected in the distribution by speech parts (POS) of the explainability scores calculated for CLIP, as shown in the App.~\ref{app:pos}. 

To alleviate this shortcoming of providing spatial positioning with textual description, we add spatial conditioning as an additional input. As far as we can ascertain, CLIP has not been used before for image generation conditioned on spatial masks. The somewhat related task of CLIP-based zero-shot image inpainting was recently successfully performed by \cite{Avrahami2021blendeddiffusion,bau2021paintbyword}, who point out that a simple masking of the input image presented to CLIP, such that the similarity score is only predicted based on a specific region of the image, does not guarantee that the generated object will not deviate from the provided region. 

{Preventing the generator from producing objects outside the designated spatial locations requires applying additional losses on the background or restricting the optimization process such that only the parts of the latent vector that affect the desired region {of the image} are optimized.} These methods limit the spatial conditioning to applications that receive an input image to be used as unaltered background.
However, since the explainability maps produced for CLIP indicate the location of the objects, we can effectively limit the location of generated objects using explainability.
\begin{algorithm}[t!]
\caption{Obtain IoU loss from masks and image.}\label{euclid}
$\mathbf{Input:}$  (i) $m_1,...,m_k$- bounding boxes of the objects to be generated, (ii)  $t_1,...,t_k$- textual descriptions of the objects we wish to generate, (iii) $C$- a pre-trained CLIP model. (iv) the input image $i$, (v) a threshold $T$ over relevancy scores (vi) $temp$ - a temperature for the sigmoid operation (vii) $expl$ - the image explainability algorithm for CLIP, which outputs relevance scores for the image patches for each pair of image and text.\\
$\mathbf{Output:}$ $\mathcal{L}_{IoU}$- an explainability-based IoU loss for the input masks $m_1,...,m_k$, and input image $i$.
\begin{algorithmic}[1]
\STATE $\mathcal{L}_{IoU} \gets 0$
\STATE \emph{for} $j\in \{1,\dots,k\}$,:
\STATE ~~~~~~$R_j \gets expl(i, t_j)$
\STATE ~~~~~~$R_j \gets R_j / R_j.max()$
\STATE ~~~~~~$pred\_mask_j \gets sigmoid((R_j - T) * temp)$ 
\STATE ~~~~~~$intersection \gets \sum_{p\in i}(pred\_mask_j[p] \cdot m_j[p])$
\STATE ~~~~~~$\mathcal{L}_{IoU} \gets \mathcal{L}_{IoU} + \frac{2*intersection}{\sum_{p\in i}pred\_mask_j[p]+\sum_{p\in i}m_j[p]}$

\end{algorithmic}
\label{alg:segm}
\end{algorithm}

Our method employs an IoU-inspired loss based on the explainability of the image modality. Alg.~\ref{alg:segm} describes how we produce the loss  $\mathcal{L}_{IoU}$ given the input spatial conditioning masks $m_1,...,m_k$ and the input image $i$. For each bounding box $m_j$ and the text $t_j$ describing the object we wish to generate in that location, we generate the explainability for CLIP with the entire image $i$ and text $t_j$ (L.3). This explainability map represents the location in which the object is currently found in the image by CLIP. 
We then transform the explainability map into a semi-binary mask (L.5) by substracting a threshold value $T$ and passing the output through a sigmoid function with high temperature $temp$. This predicted mask is then used to calculate a Dice Loss with respect to the ground truth object mask (L.7). After calculating the IoU-based loss, we incorporate the similarity-based loss $     \mathcal{L} = -\lambda_{expl}\cdot\mathcal{L}_{IoU} - \sum_{j=1}^k\text{CLIP}(i, t_j)$,
where $\lambda_{expl}$ is a hyperparameter, and the sum calculates the CLIP similarity between the image and each object we wish to generate, in order to ensure that all objects in $\{t_1,...,t_k\}$ appear in $i$. $\lambda_{expl}$, $T$ and $temp$ are chosen empirically, using examples from the MSCOCO~\cite{ty2014coco} validation set.

\section{Experiments}

We evaluate our method in various contexts, including one-shot prompt engineering for image classification based on~\cite{zhou2021coop}, {zero-shot} text-guided image manipulation based on~\cite{patashnik2021styleclip}, and {zero-shot} text-guided image generation with spatial conditioning based on CLIP-guidance for VQGAN~\cite{Esser_2021_CVPR,vqgan+clip}.

\subsection{One-shot prompt engineering}
\label{sec:prompt_exp}

\begin{figure*}[t!]
  \centering
  \begin{tabular}{c@{~~}c@{~~~~}c@{~~~~}c@{~~~~}c@{~~~~}c}
   & \begin{small} Original \end{small}  & \multicolumn{2}{c}{\begin{small} Ground truth class label \end{small}} & \multicolumn{2}{c}{\begin{small} Counter factual class label \end{small}} \\
   & \begin{small} image \end{small} & \begin{small} CoOp~\cite{zhou2021coop} \end{small} & \begin{small} Ours \end{small} & \begin{small} CoOp~\cite{zhou2021coop} \end{small} & \begin{small} Ours \end{small}  \\

&
&
\includegraphics[width=0.17\linewidth,clip] {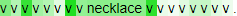} &
\includegraphics[width=0.17\linewidth,clip] {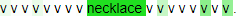}&
\includegraphics[width=0.17\linewidth,clip] {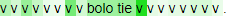} &
\includegraphics[width=0.17\linewidth,clip] {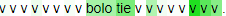} \\

&
\includegraphics[width=0.14\linewidth,clip] {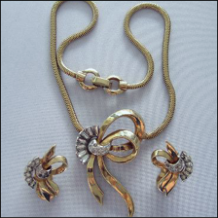} &
\includegraphics[width=0.14\linewidth,clip] {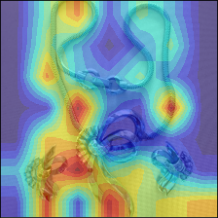}&
\includegraphics[width=0.14\linewidth, clip]{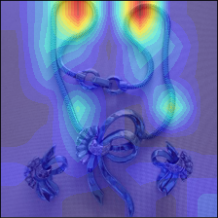}&
\includegraphics[width=0.14\linewidth,clip] {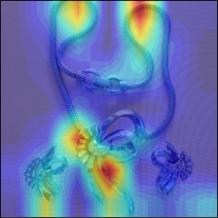} &
\includegraphics[width=0.14\linewidth,clip] {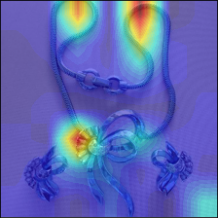}\\

&
&
\includegraphics[width=0.17\linewidth,clip] {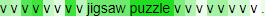} &
\includegraphics[width=0.17\linewidth,clip] {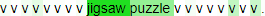}&
\includegraphics[width=0.17\linewidth,clip] {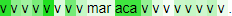}&
\includegraphics[width=0.17\linewidth,clip] {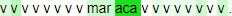}\\

&
\includegraphics[width=0.14\linewidth, clip]{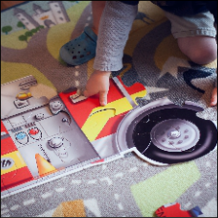}&
\includegraphics[width=0.14\linewidth,clip] {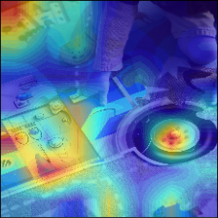}&
\includegraphics[width=0.14\linewidth, clip]{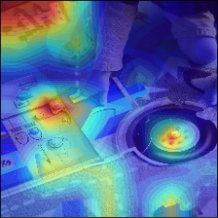}&
\includegraphics[width=0.14\linewidth,clip] {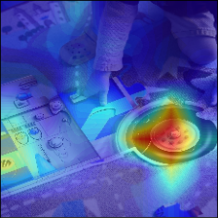} &
\includegraphics[width=0.14\linewidth,clip] {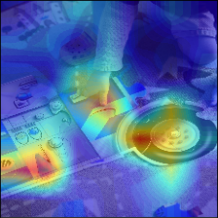}\\

\end{tabular}
\caption{A qualitative comparison of prompt engineering using CoOp~\cite{zhou2021coop} with and without our method on 2 exemplary samples from ImageNetV2~\cite{pmlr-v97-recht19a}. We present the relevance maps for the ground truth class chosen by our method (``necklace", ``jigsaw"), and the counterfactual class chosen by CoOp (``bolo tie", ``maraca"). The learned vectors for the prompt are annotated by the letter ``v" in the textual explainability maps, since the vectors do not represent actual tokens. As can be seen, for the ground truth classes ``necklace" and ``jigsaw", our prompts encourage CLIP to focus on the class name in the input text, while CoOp leads CLIP to consider unrelated tokens.
This can cause CLIP to produce biased similarity scores based on the engineered prompts. \label{fig:prompt-eng}}
\end{figure*}

\begin{table}[t]
    \centering
    \resizebox{\linewidth}{!}{
    \begin{tabular}{@{}l@{~}l@{~}c@{~}c@{~}c@{~}c@{~}c@{~}c@{~}c@{~}c@{~}c@{~}c@{}}
        \toprule
        {Image} &&\multicolumn{2}{c}{ImageNet} & \multicolumn{2}{c}{ImageNetV2} & \multicolumn{2}{c}{INet-Sketch}& \multicolumn{2}{c}{ImageNet-A} & \multicolumn{2}{c}{ImageNet-R} \\
        {backbone} && Unified  & CSC &Unified  & CSC &Unified  & CSC &Unified  & CSC &Unified  & CSC  \\
        \midrule
        \multirow{6}{*}{ResNet-50}&0-shot & 58.18  & -  & 51.34  & - & \textbf{33.32} & - & 21.65 & - & 56.00 &  -\\
        & LP & 21.70 &- & 17.78 & - & 5.57 & - & 0.11 & - & 0.07 & -\\
        &CoOp M=16& 54.45 & 28.40  & 47.11& 23.92 & 28.12 & 11.80 & 19.97 & 10.39 & 50.38 & 26.83\\
        &CoOp M=4 & 57.63  & 35.88  & 50.34  & 30.49 & 30.18 & 16.28 & 21.43 & 13.45 & 53.53 &  32.06\\
         &Ours M=16 & 58.13 & 31.90 & 51.30 & 26.82 & 32.49 & 13.52 & 22.12& 11.77 & \textbf{57.73}& 29.26 \\
         &Ours M=4 & \textbf{59.05} & \textbf{38.79} & \textbf{52.33} & \textbf{33.58} & 32.59 & \textbf{18.25} & \textbf{22.74}& \textbf{14.33} & 57.15& \textbf{34.63}  \\
        \midrule
        \multirow{6}{*}{ResNet-101} &0-shot & 61.62  & -  & 54.81  & - & \textbf{38.71} & - & 28.05 & - & 64.38 &  -\\
        & LP & 26.41 & - & 21.75 & - & 9.61 & - & 0.08 & - & 0.07 & -\\
        &CoOp M=16 & 57.84  & 33.51  & 51.25  & 26.98 & 33.80 & 15.78 & 26.82 & 14.28 & 59.02 &  32.40\\
        &CoOp M=4 & 60.41  & 38.96  & 53.68  & 33.43 & 36.71 & 21.19 & 27.94 & 16.91 & 61.08 &  40.27\\
        &Ours M=16 & 61.76 &36.01 & 55.02 & 31.07 & 37.96 & 18.70 & 29.56& 15.97 & 63.92& 36.02 \\
        &Ours M=4 & \textbf{62.31} & \textbf{40.77} & \textbf{55.65} & \textbf{35.18} & 38.51 & \textbf{21.88} & \textbf{30.07}& \textbf{17.80} & \textbf{65.33}& \textbf{40.44}  \\
        \midrule
        \multirow{6}{*}{ViT-B/16} &0-shot & 66.73  & -  & 60.83  & - & 46.15 & - & 47.77 & - & 73.76 &  -\\
        & LP & 32.26 & - & 27.33 & - & 16.48 & - & 0.10 & - & 0.08 & -\\
        &CoOp M=16 & 63.66  &  38.86 &  56.53 & 33.55 & 40.96 & 22.59 & 43.93& 23.30 & 69.33& 42.76 \\
        &CoOp M=4 & 66.93  & 46.20  & 60.14  & 40.28 & 44.97 & 28.26& 47.44 & 31.87 & 72.12 &  51.16\\
        &Ours M=16 & 67.09& 40.78 & 60.28 & 35.25 &45.71 & 23.77& 48.29& 25.03 &74.9& 44.43 \\
        &Ours M=4 & \textbf{67.62} & \textbf{48.74} & \textbf{61.07} & \textbf{42.58} & \textbf{46.33} & \textbf{30.34} & \textbf{49.46}& \textbf{34.08} & \textbf{75.66}& \textbf{53.75}  \\
        \midrule
        \multirow{6}{*}{ViT-B/32} &0-shot & 62.05  & -  & 54.79  & - & \textbf{40.82} & - & 29.57 & - & 65.99 &  -\\
        & LP & 27.03 & - & 22.38 & - & 11.32 & - & 0.12 & - &  0.08 & -\\
        &CoOp M=16 &57.64  &  33.42 &50.24 & 28.39 & 35.12 & 17.63& 27.53 &  13.84 & 59.46& 34.30 \\
        &CoOp M=4 & 61.48  & 40.66  & 54.01  & 34.52 & 38.26 & 22.76 & 29.56 & 18.58 & 63.11&  41.12\\
        &Ours M=16 & 62.55 & 38.63 & 55.14 & 33.23 & 40.40& 21.08& 31.22& 16.8 & \textbf{67.22}& 39.64 \\
        &Ours M=4 & \textbf{63.69} & \textbf{42.98} & \textbf{55.84} & \textbf{37.21} & 40.23 & \textbf{24.26} & \textbf{30.78}& \textbf{20.48} & 66.49& \textbf{44.22}  \\
        \bottomrule
    \end{tabular}
    }
    \smallskip
    \smallskip
    \caption{1-shot accuracy (in percentage) of  linear probing (LP) and CLIP~\cite{radford2021learning} with prompts produced by the method of \cite{zhou2021coop} (CoOp) or with our explainability-guided variant, with various image backbones. All methods are trained on ImageNet and evaluated on several variants. Unified stands for training a single prompt for all classes, and CSC (class-specific) stands for optimizing a prompt for each class name. Results are averaged over 3 random seeds.}
    \label{table:promptengineering}
\end{table}

We compare the classification accuracy of CLIP using the prompts optimized with CoOp~\cite{zhou2021coop} and with our method, as described in Sec.~\ref{sec:prompt}. Following~\cite{zhou2021coop}, we evaluate the methods on 
ImageNet~\cite{Deng2009ImageNetAL} test set,
ImageNetV2~\cite{pmlr-v97-recht19a},  
ImageNet-Sketch~\cite{NEURIPS2019_3eefceb8}, 
ImageNet-A~\cite{Hendrycks_2021_CVPR}, 
and Imagenet-R~\cite{Hendrycks_2021_ICCV}.

Following~\cite{zhou2021coop}, two scenarios are tested: unified prompt engineering and class-specific prompt engineering. In the unified scenario, a single prompt is optimized for all class names. In the class-specific (CSC) case, a different prompt is optimized per class. Note that for all datasets, the prompts are optimized using labeled examples only from the ImageNet training set, in order to test the robustness of the optimized prompts on different ImageNet variations. 

For both methods we test different backbones for the visual encoder of CLIP (see Tab.~\ref{table:promptengineering}), including variations of ViT~\cite{dosovitskiy2020image} and of ResNet~\cite{he2016deep}.
Following~\cite{zhou2021coop}, we optimize a template with $M=16$ tokens. We also include results for $M=4$, as it was noted to sometimes achieve superior results on ImageNet.

Two options for positioning the class name tokens in the prompt were reported in~\cite{zhou2021coop}, with similar outcomes. The first has the class name located in the middle of the prompt, i.e.: $t=v_1,...,v_8, \{\text{label}\}, v_9,...,v_{16},$ where $v_1,...,v_{16}$ are the prompt tokens, and the second has the class name located at the end, i.e.: $t=v_1,...,v_{16},\{\text{label}\}$. In the main text we report the results of the former; for the latter, see App.~\ref{app:prompt-eng-end}.
We use $\lambda_{expl}=1$ for experiments that use ViT-B/16 as backbone and $\lambda_{expl}=3$ for all other backbones. \\
Tab.~\ref{table:promptengineering} shows the 1-shot accuracy of CoOp and our method, in addition to 0-shot manual prompt selection and linear probing of the image embedding produced by CLIP, which are the baselines used by CoOp~\cite{zhou2021coop}. 2-shot and 4-shot results are available in App.~\ref{app:prompt-eng-few-shot}.
As can be seen,  both linear probing and CoOp are heavily overfitting and actually achieve significantly lower accuracy than 0-shot results.
Using the explainability-based loss, our method is consistently able to improve upon CoOp,
leading to higher accuracy across all backbones, all datasets, and both scenarios (unified and CSC).

A Sensitivity analysis for $\lambda_{expl}$ is presented in Fig. ~\ref{fig:PE_lambda_sensitivity}, showing that the improvement in accuracy is consistent across a large range of $\lambda_{expl}$ values. Fig.~\ref{fig:prompt-eng} presents a qualitative comparison of using CoOp with and without our method, see caption for a detailed description.
\subsection{Zero-shot text-guided image manipulation}

Next, we compare our explainability-based optimization (Sec.~\ref{sec:manipulation}) with the optimization presented in~\cite{patashnik2021styleclip}. There are three methods for text-based image editing using StyleGAN~\cite{karras2020analyzing} presented by~\cite{patashnik2021styleclip} -  latent optimization, mapper training, and global directions extraction. We focus on latent optimization, since our focus is on zero-shot methods and the other two methods employ additional training.
As described in Sec.~\ref{sec:manipulation}, we add the explainability-based loss from Eq.~\ref{eq:expl_gen_loss} to the optimization algorithm of~\cite{patashnik2021styleclip}. We choose the set of hyperparameters for our explainability-based loss from the set: $\lambda_{expl} = \{0, 0.5, 1, 1.5, 2, 2.5, 3, 3.5\}$, and use the best value for $\lambda_{expl}$ according to the highest CLIP similarity score. 

Since the optimization in~\cite{patashnik2021styleclip} requires a different hyperparameter setting for each prompt, we select prompts that appear in the paper or official code, and use the same hyperparameters (in other words, we do not manually select the hyperparameters for our method). Next, we choose 20 random seeds to be used across all our experiments to generate the images to be edited, $i$. For each image $i$, and text prompt $t$ we produce the edited image with StyleCLIP's optimization, and with our modified optimization. 

For evaluation, we extract all examples where our method produces a different output than StyleCLIP, i.e., all cases where the automatic procedure selected $\lambda_{expl}\neq 0$, and conduct a user study among $46$ users. Users were asked to evaluate each manipulation by the similarity between the manipulated image and the input prompt $t$ and by the loss of identity between the manipulated image and the original image $i$, both on a scale of $1-5$ (higher is better). 

\begin{figure*}[t!]
  \centering
  
   \begin{tabular}{c@{~}c@{~}c@{~~~}c@{~}c@{~~~}c@{~}c@{~~~}c@{~}c}
\centering{\begin{small}\begin{turn}{90} Original \end{turn}\end{small}}
&
\includegraphics[width=0.10\linewidth, clip]{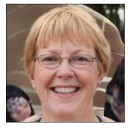}&
\includegraphics[width=0.10\linewidth, clip]{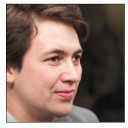}&
\includegraphics[width=0.10\linewidth, clip]{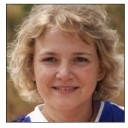}&
\includegraphics[width=0.10\linewidth, clip]{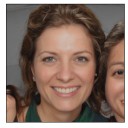}&
\includegraphics[width=0.10\linewidth, clip]{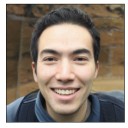} &
\includegraphics[width=0.10\linewidth, clip]{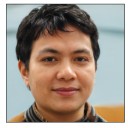} &
\includegraphics[width=0.10\linewidth, clip]{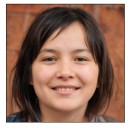} &
\includegraphics[width=0.10\linewidth, clip]{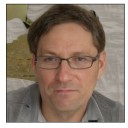}\\
\centering{\begin{small}\begin{turn}{90} ~~SC \end{turn}\end{small}}
&
\includegraphics[width=0.10\linewidth, clip]{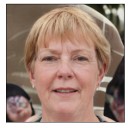}&
\includegraphics[width=0.10\linewidth, clip]{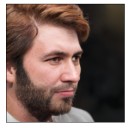}&

\includegraphics[width=0.10\linewidth, clip]{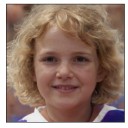}&
\includegraphics[width=0.10\linewidth, clip]{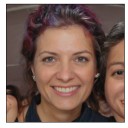}&

\includegraphics[width=0.10\linewidth, clip]{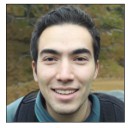} &
\includegraphics[width=0.10\linewidth, clip]{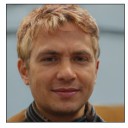} &

\includegraphics[width=0.10\linewidth, clip]{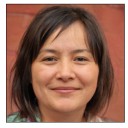}
&
\includegraphics[width=0.10\linewidth, clip]{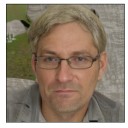}\\
\centering{\begin{small}\begin{turn}{90} ~Ours \end{turn}\end{small}}
&
\includegraphics[width=0.10\linewidth, clip]{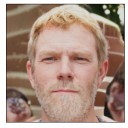}&
\includegraphics[width=0.10\linewidth, clip]{user_study/beard/ours_sc_2.jpg}&

\includegraphics[width=0.10\linewidth, clip]{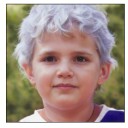}&
\includegraphics[width=0.10\linewidth, clip]{user_study/purple/ours_sc_2.jpg}&

\includegraphics[width=0.10\linewidth, clip]{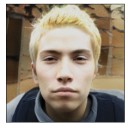} &
\includegraphics[width=0.10\linewidth, clip]{user_study/blond/ours_sc_2.jpg} &

\includegraphics[width=0.10\linewidth, clip]{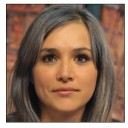}
&

\includegraphics[width=0.10\linewidth, clip]{user_study/grey/ours_sc_2.jpg}\\
\cmidrule(r){2-3}
\cmidrule(r){4-5}
\cmidrule(r){6-7}
\cmidrule(r){8-9}
&\multicolumn{2}{c}{(a)}~~ & \multicolumn{2}{c}{(b)}~ & \multicolumn{2}{c}{(c)}~ & \multicolumn{2}{c}{(d)}
\end{tabular}
\caption{A qualitative comparison between StyleCLIP {(SC)} and our method on 4 different textual prompts. (a) ``A man with a beard'', (b) ``A person with purple hair'', (c) ``A blond man'', (d) ``A person with grey hair''. For each prompt we present examples where StyleCLIP is successful (right column), and unsuccessful (left column). For the failure cases, the optimization in StyleCLIP hardly modifies the original image, leading to a high identity preservation score when no semantic change was applied. When StyleCLIP is successful, our method produces similar or identical results. \label{fig:userstudy-examples}}
\end{figure*}

\begin{table*}[t!]
\centering
\resizebox{1.0\linewidth}{!}{
    \begin{tabular}{lccccccccc}
    \toprule
    & \multicolumn{2}{c}{A man with a beard} & \multicolumn{2}{c}{A person with purple hair}& \multicolumn{2}{c}{A blond man}& \multicolumn{2}{c}{A person with grey hair}\\
    \cmidrule(lr){2-3}
    \cmidrule(lr){4-5}
    \cmidrule(lr){6-7}
    \cmidrule(lr){8-9}
    Method &Quality&Identity &Quality&Identity &Quality&Identity &Quality&Identity\\
    \midrule
    SC & 2.92 & \textbf{3.61} & 1.17 & \textbf{4.13} & 3.93 & \textbf{3.67}& 2.59 & \textbf{3.82} \\
    Ours & \textbf{4.28} & 2.23 & \textbf{2.29} & 3.51 & \textbf{4.28}& 2.63 & \textbf{3.27}& 3.10\\    
    \bottomrule
    \end{tabular}
    }\smallskip \smallskip
    \caption{A user study comparing text-based image editing with StyleCLIP (SC) and our method on 4 different textual prompts: ``A man with a beard'', ``A person with purple hair'', ``A blond man'', ``A person with grey hair''. Quality refers to the similarity between the prompt and the manipulation; Identity refers to the identity preservation of the manipulation.  Scores are averaged across 20 random seeds, on a scale of 1-5 (higher is better).}
    \label{fig:userstudy}
\end{table*}

Fig.~\ref{fig:userstudy-examples} presents sample results from our user study (See App.~\ref{app:sc} for full results). Notice that for challenging manipulations, such as using the prompt ``a man with a beard'' on a woman, StyleCLIP tends to leave the input image $i$ almost unchanged. In contrast, in many of these cases, our method compels the optimization to reach a solution that fits the required semantic edit. We present the results of the user study for each prompt in Tab.~\ref{fig:userstudy} (see results with standard deviation in App.~\ref{app:sc}). As can be seen, our method produces results that are consistently rated by users as more similar to the target text. However, StyleCLIP, which, as can be seen in Fig.~\ref{fig:userstudy-examples}, often leaves the images unchanged, obtains a higher identity preservation score. Evidently, the gap in the identity score is much bigger for the prompts ``A \textbf{man} with a beard'' and ``A blond \textbf{man}''. These prompts modify the gender of the subject of the image $i$, thereby requiring a more substantial identity change. 

\subsection{Zero-shot text-to-image with spatial conditioning}

\begin{figure*}[t!]
  \centering
  \begin{tabular}{ccccc}
   Input & Textual &Similarity- & Similarity- & ours \\
   conditioning & conditioning & based & based 2 &\\
   
\includegraphics[width=0.17\linewidth,clip]
{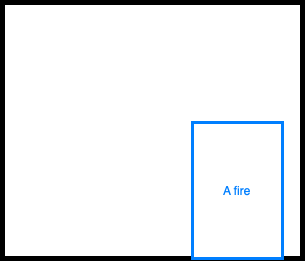}&
\includegraphics[width=0.17\linewidth,clip]
{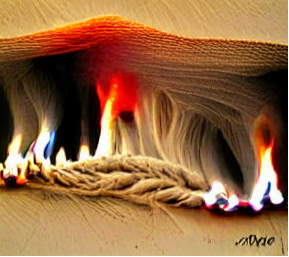}&
\includegraphics[width=0.17\linewidth,clip]
{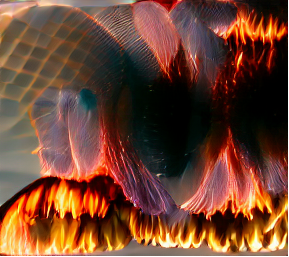}&
\includegraphics[width=0.17\linewidth,clip]
{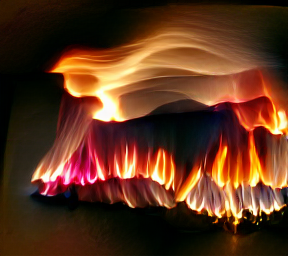}&
\includegraphics[width=0.17\linewidth,clip]
{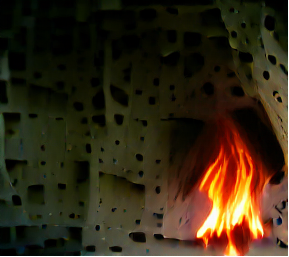}\\

\includegraphics[width=0.17\linewidth, clip]
{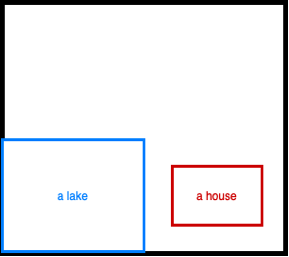}&
\includegraphics[width=0.17\linewidth, clip]
{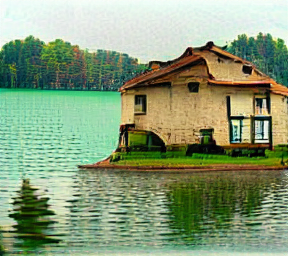}&
\includegraphics[width=0.17\linewidth,clip] 
{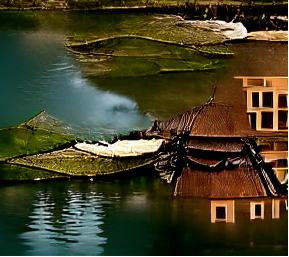}&
\includegraphics[width=0.17\linewidth,clip] 
{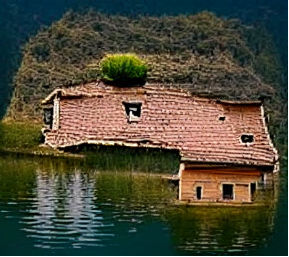}&
\includegraphics[width=0.17\linewidth,clip] 
{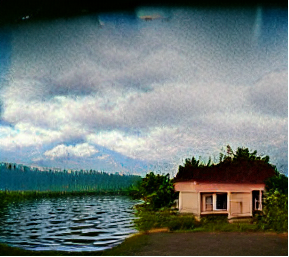}\\

\includegraphics[width=0.17\linewidth,clip]
{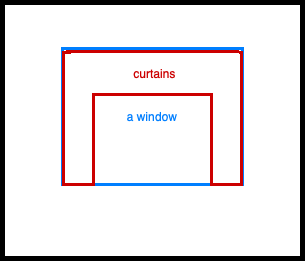}&
\includegraphics[width=0.17\linewidth,clip]
{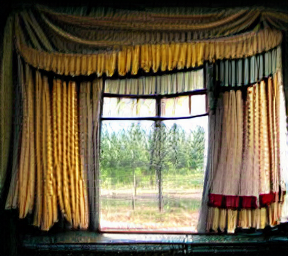}&
\includegraphics[width=0.17\linewidth,clip]
{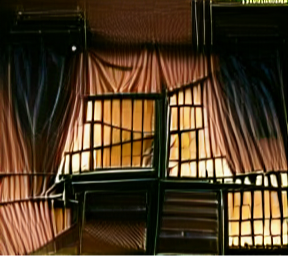}&
\includegraphics[width=0.17\linewidth,clip]
{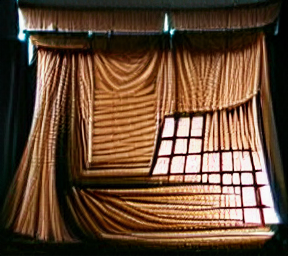}&
\includegraphics[width=0.17\linewidth,clip]
{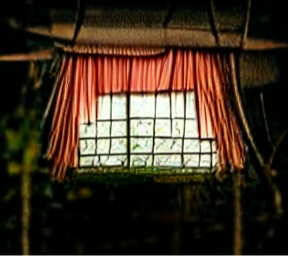}\\

\includegraphics[width=0.17\linewidth,clip]
{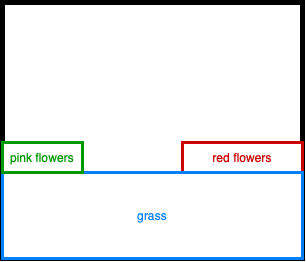}&
\includegraphics[width=0.17\linewidth,clip]
{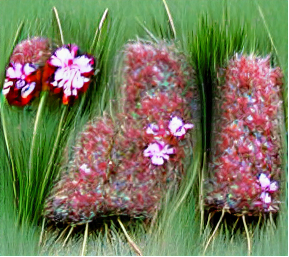}&
\includegraphics[width=0.17\linewidth,clip]
{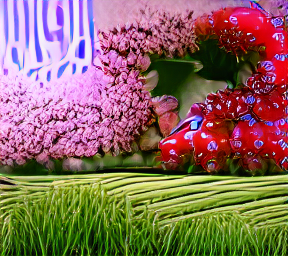}&
\includegraphics[width=0.17\linewidth,clip]
{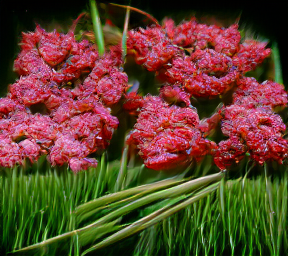}&
\includegraphics[width=0.17\linewidth,clip]
{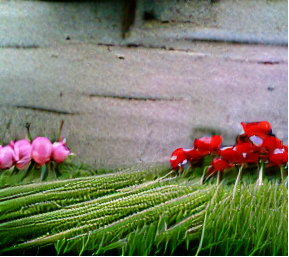}\\

\includegraphics[width=0.17\linewidth,clip] 
{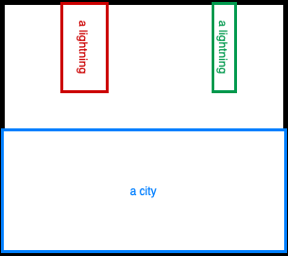}&
\includegraphics[width=0.17\linewidth,clip] 
{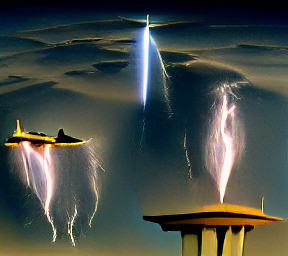} &
\includegraphics[width=0.17\linewidth,clip] 
{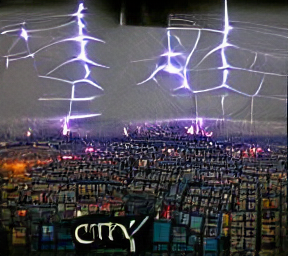} &
\includegraphics[width=0.17\linewidth,clip] 
{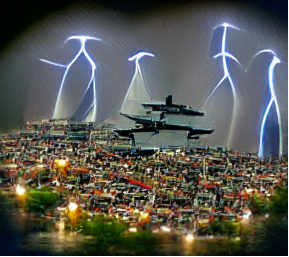}&
\includegraphics[width=0.17\linewidth,clip] 
{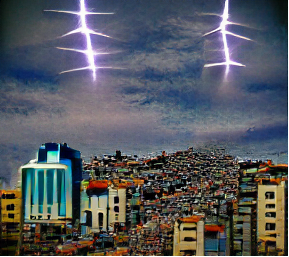}\\

\includegraphics[width=0.17\linewidth,clip] 
{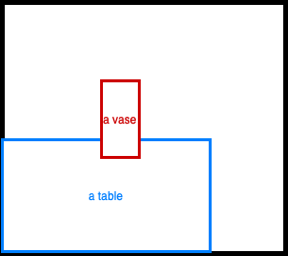}&
\includegraphics[width=0.17\linewidth,clip] 
{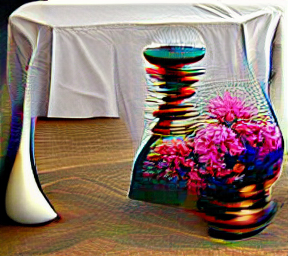}&
\includegraphics[width=0.17\linewidth,clip] 
{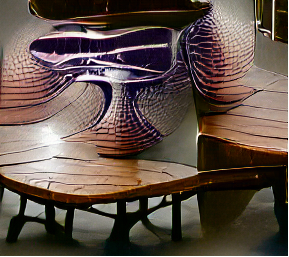}&
\includegraphics[width=0.17\linewidth,clip] 
{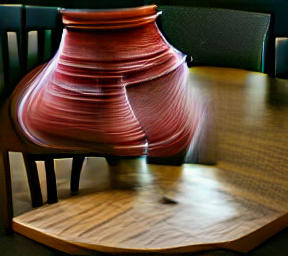}&
\includegraphics[width=0.17\linewidth,clip] 
{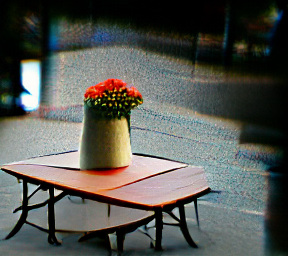}\\
\end{tabular}
\caption{A qualitative comparison between the two similarity-based baselines and our method for CLIP-guided zero-shot text-based image generation with spatial conditioning. {Textual conditioning refers to specifying the spatial positioning of objects within the text prompts, for example "a vase on a table". Additional examples are presented in App.~\ref{app:l2i_examples}.\label{fig:l2i}}}
\end{figure*}

We use CLIP-guided VQGAN as implemented by~\cite{vqgan+clip}. Since, as far as we can ascertain, there is no previous literature on zero-shot CLIP-guided text-to-image generation with spatial conditioning on the location of the generated objects, we use two variations of a similarity-based CLIP loss to create  baselines without explainability conditioning. The first baseline employs the loss $    \mathcal{L}_{masked} = \sum_{t\in \{t_1,...,t_k\}}\sum_{m\in \{m_1,...,m_k\}}- \text{CLIP}(i_m, t),$
where $i_m$ is the image $i$ masked according to bounding box $m$, i.e. for each mask $m$, we black out all pixels outside $m$, in order to ensure that the objects identified by CLIP reside within the bounding boxes, and $t$ is a prompt of the form ``a photo of \{label\}'' where "label" is the target class to be generated in bounding box $m$. This masking technique has also been employed in previous works~\cite{Avrahami2021blendeddiffusion,bau2021paintbyword}, for CLIP-guided image inpainting.
The second similarity-based baseline we consider employs the loss $\mathcal{L}_{masked}$ in addition to the similarity loss in the unmasked image $     \mathcal{L}_{similarity} = \sum_{t\in \{t_1,...,t_k\}}-\text{CLIP}(i, t)$. 
This baseline uses the loss: $\mathcal{L} = \mathcal{L}_{similarity} + \mathcal{L}_{masked}$, which considers both the information inside the bounding boxes and the information in the entire image.

As mentioned in Sec.~\ref{sec:l2i}, since a simple similarity-based loss has no spatial restrictions, the baselines produce objects outside the input bounding box (see Fig.~\ref{fig:l2i}), while our loss produces objects within the bounding box, thanks to spatial conditioning based on explainability. Moreover, the examples in Fig.~\ref{fig:l2i} demonstrate the ability of our method to generate images in a variety of cases, including multiple bounding boxes with varying heights and widths (see App.~\ref{app:l2i_examples} for additional examples and visualizations of the explainability maps).
As smaller bounding boxes require stronger supervision, we set $\lambda_{expl_i}$ for object $i$ to be $\lambda_{expl_i}=\frac{0.15}{\sqrt{r(m_i)}}$, where $m_i$ is the bounding box assigned to object $i$ and $r(m_i)$ is the ratio between the area of the mask and the area of the entire image. 
The threshold $T$ is set to 0.1 and $temp$ is set to 20.

\begin{table}[t]
    \centering
      \begin{minipage}[c]{0.49\linewidth}
  \begin{tabular*}{\linewidth}{@{\extracolsep{\fill}}lccc}
        \toprule
        & Similarity- & Similarity- & Ours \\
        &based & based 2&\\
        \midrule
        Precision & 46.4 & 26.9 & \textbf{71.7}\\
        Recall &  48.3 & 30.5 & \textbf{63.4}\\
        F1 & 40.5 & 24.28 &\textbf{62.6} \\
        AP & 8 & 5.4 & \textbf{26.2}\\
        AR &  21.6 & 19 & \textbf{40}\\
        AP$_{0.5}$ & 18 & 15.4 & \textbf{56.5} \\
        \bottomrule
    \end{tabular*}
        \captionof{table}{Precision, recall, F1, average precision, and average recall for spatially conditioned image generation with our method, and two similarity-based baselines (results in percentage). Metrics were averaged across $100$ random samples from the MSCOCO~\cite{ty2014coco} validation set and four random seeds. Average precision and average recall are calculated using DETR~\cite{carion2020end}.}
    \label{tab:l2i_metrics}
\end{minipage}%
\hfill
 \begin{minipage}[c]{0.49\linewidth}
\includegraphics[width=\linewidth, clip]
{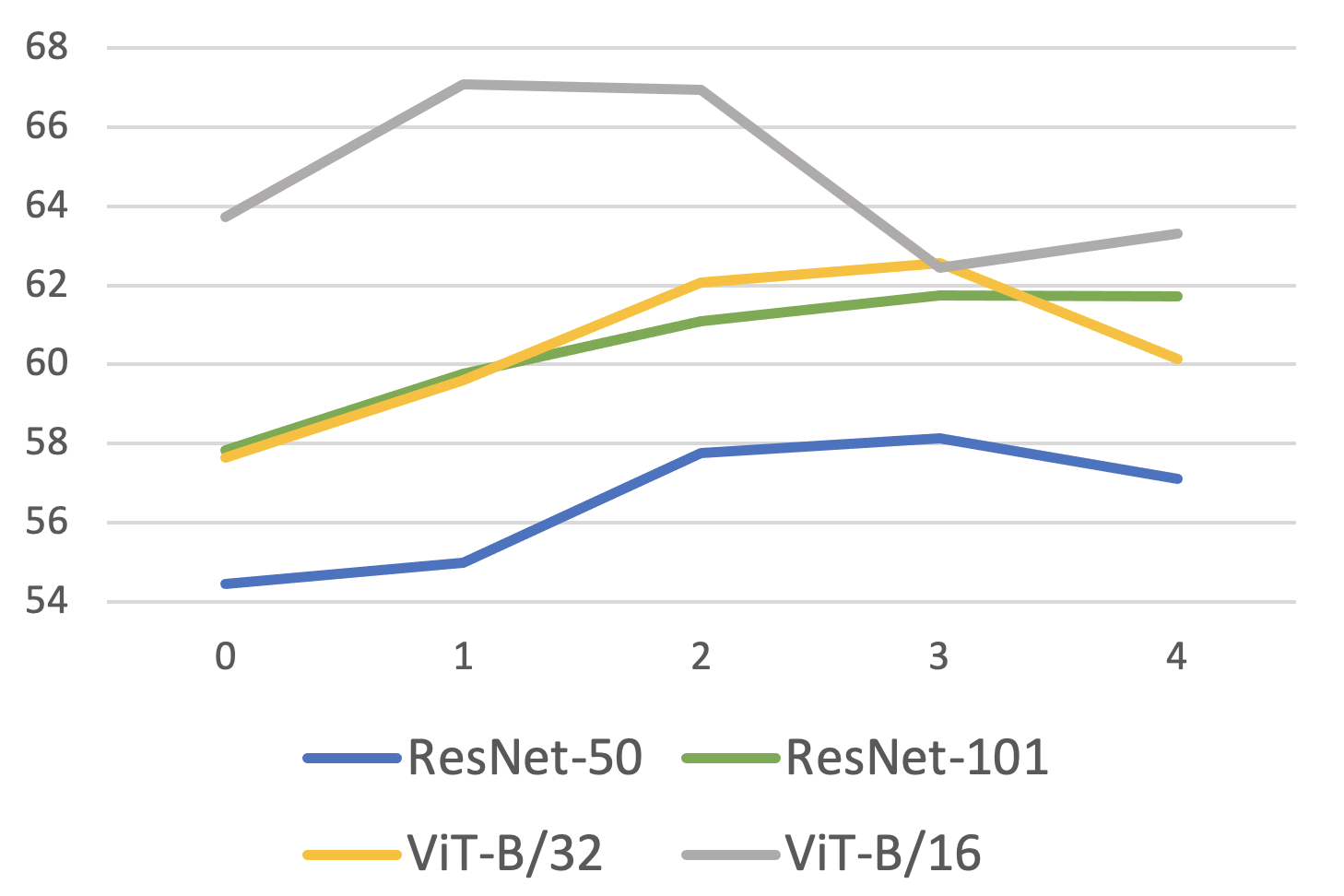}\\
\captionof{figure}{1-shot accuracy (in percentage) on the ImageNet test set for different choices of $\lambda_{expl}$ for all visual backbones of CLIP. The accuracy achieved by the baselines is denoted as $\lambda_{expl}=0$.} \label{fig:PE_lambda_sensitivity}
\end{minipage}%

\end{table}

In order to provide quantitative metrics for our spatially conditioned text-to-image generation, we use the validation set from MSCOCO~\cite{ty2014coco} which contains bounding boxes for each object in the image. In order to ensure a varying number and size of objects, while maintaining enough background to allow object-free generation, which is challenging for CLIP-guided VQGAN, we filter the layout as follows: we keep the $k$ largest bounding boxes whose commutative area is less than 50\% of the image, where adding the next largest bounding box would result in occupying more than 50\% of the image. By focusing on the largest objects we also help ensure that the size of each bounding box suffices for the CLIP encoder. 
For our first experiment, we sample 100 MSCOCO images at random, and use our method and the similarity-based baselines to generate images corresponding to the annotated spatial layout. We then produce an explainability map for each text description $t$, as described in Alg.~\ref{alg:segm} (L.2-3). We use these maps as soft semantic segmentation, binarize them using thresholds produced with Otsu's method~\cite{otsu1979threshold}, and calculate the precision, recall, and F1 scores of the binarized maps with the ground truth bounding boxes $m_1,.., m_k$.  Note that both precision and recall are limited and cannot reach $100\%$ due to the square shape of the bounding boxes, which is not suited to non-square objects. Precision is also limited because images often contain more than one instance of a specific class, leading to a high explainability score for the other occurrences as well.
As can be seen in Tab.~\ref{tab:l2i_metrics}, our method significantly outperforms the baselines. \\
Next, we use object detection to evaluate the quality of the generated objects, as well as the overlap between their location and the target spatial condition. DETR~\cite{carion2020end} is used to produce bounding boxes for each object. These bounding boxes are evaluated against the input spatial conditioning masks using the average precision and average recall scores. As can be seen in Tab.~\ref{tab:l2i_metrics}, our method greatly outperforms the baselines in this evaluation as well, implying that the explainability signal is indeed indicative enough to enforce spatial restrictions over an image.

\section{Discussion}

In our experiments, we presented a generic application of explainability to improve classification, image editing, and image synthesis. There are also specific situations in which a limited view of the input is detrimental and where explainability can help ensure a more even distribution of information pooling. One such case, studied in App.~\ref{app:fusedream}, is that of compound nouns, e.g. ``apple juice'' or ``blueberry muffin''. As we show, state-of-the-art {zero-shot} text-to-image generation engines might overly emphasize or ignore some of the textual input, leading to misinterpretation of the text. The method we present for equalizing the contributions {to avoid such neglect} not only leads to considerably better image outputs for such cases, but also slightly improves the FID score for other sentences. See App.~\ref{app:fusedream} for full details of the method implementation, visual examples, and the results of a user study conducted against results obtained with a state-of-the-art method. In order to demonstrate the wide applicability of our approach, we have modified multiple zero-shot and one-shot approaches. While the baseline approaches are impressive, we do note that they are not yet ready to replace supervised methods. Prompt engineering is not yet competitive with supervised methods, CLIP-guided VQGAN often generates substandard images, and StyleCLIP optimization method often fails and requires different parameters for each prompt. Therefore, other signals need to be considered to allow zero-shot applications to compete against fully supervised ones. Explainability, as we show, is an example of such beneficial signal.

\section{Conclusions}

While explainability methods are constantly improving, their use as a feedback mechanism to improve classification or generation methods is still relatively unexplored. As far as we can ascertain, their utilization as such a building block is currently limited to weakly supervised segmentation~\cite{Zabari2021SemanticSI,Chefer_2021_ICCV}. In this work, we show how explainability can help overcome the neglect problem of bi-modal transformers, which have become a cornerstone in the current rapid evolution of zero-shot methods. We demonstrate how preventing neglect, as reflected through the lens of the explainability score, helps improve one-shot classification, zero-shot image editing, and zero-shot layout-conditioned image generation. In the first two domains, neglect is prevented in the text domain, while in the latter, the constraint on the heatmap is placed in the image domain. 

\subsubsection{Acknowledgments}
This project has received funding from the European Research Council (ERC) under the European Unions Horizon 2020 research and innovation programme (grant ERC CoG 725974). We thank Ariel Landau for his assistance.

\bibliographystyle{splncs04}
\bibliography{xai}

\begin{thebibliography}{10}
\providecommand{\url}[1]{\texttt{#1}}
\providecommand{\urlprefix}{URL }
\providecommand{\doi}[1]{https://doi.org/#1}

\bibitem{abnar2020quantifying}
Abnar, S., Zuidema, W.: Quantifying attention flow in transformers. arXiv
  preprint arXiv:2005.00928  (2020)

\bibitem{akbik2019flair}
Akbik, A., Bergmann, T., Blythe, D., Rasul, K., Schweter, S., Vollgraf, R.:
  Flair: An easy-to-use framework for state-of-the-art nlp. In: {NAACL} 2019,
  2019 Annual Conference of the North American Chapter of the Association for
  Computational Linguistics (Demonstrations). pp. 54--59 (2019)

\bibitem{akbik2018coling}
Akbik, A., Blythe, D., Vollgraf, R.: Contextual string embeddings for sequence
  labeling. In: {COLING} 2018, 27th International Conference on Computational
  Linguistics. pp. 1638--1649 (2018)

\bibitem{bau2021paintbyword}
Bau, D., Andonian, A., Cui, A., Park, Y., Jahanian, A., Oliva, A., Torralba,
  A.: Paint by word. arXiv preprint arXiv:2103.10951  (2021)

\bibitem{Berg_animals}
Berg, T., Forsyth, D.: Animals on the web. In: CVPR (2006)

\bibitem{binder2016layer}
Binder, A., Montavon, G., Lapuschkin, S., M{\"u}ller, K.R., Samek, W.:
  Layer-wise relevance propagation for neural networks with local
  renormalization layers. In: International Conference on Artificial Neural
  Networks. pp. 63--71. Springer (2016)

\bibitem{brock2018large}
Brock, A., Donahue, J., Simonyan, K.: Large scale {GAN} training for high
  fidelity natural image synthesis. In: International Conference on Learning
  Representations (2019), \url{https://openreview.net/forum?id=B1xsqj09Fm}

\bibitem{carion2020end}
Carion, N., Massa, F., Synnaeve, G., Usunier, N., Kirillov, A., Zagoruyko, S.:
  End-to-end object detection with transformers. arXiv preprint
  arXiv:2005.12872  (2020)

\bibitem{chefer2021imagebased}
Chefer, H., Benaim, S., Paiss, R., Wolf, L.: Image-based clip-guided essence
  transfer (2021)

\bibitem{Chefer_2021_ICCV}
Chefer, H., Gur, S., Wolf, L.: Generic attention-model explainability for
  interpreting bi-modal and encoder-decoder transformers. In: Proceedings of
  the IEEE/CVF International Conference on Computer Vision (ICCV). pp. 397--406
  (October 2021)

\bibitem{Chefer_2021_CVPR}
Chefer, H., Gur, S., Wolf, L.: Transformer interpretability beyond attention
  visualization. In: Proceedings of the IEEE/CVF Conference on Computer Vision
  and Pattern Recognition (CVPR). pp. 782--791 (June 2021)

\bibitem{Chen2015WeblySL}
Chen, X., Gupta, A.K.: Webly supervised learning of convolutional networks.
  2015 IEEE International Conference on Computer Vision (ICCV) pp. 1431--1439
  (2015)

\bibitem{vqgan+clip}
Crowson, K.: Vqgan+clip.
  https://colab.research.google.com/drive/1L8oL-vLJXVcRzCFbPwOoMkPKJ8-aYdPN
  (2021)

\bibitem{Deng2009ImageNetAL}
Deng, J., Dong, W., Socher, R., Li, L.J., Li, K., Fei-Fei, L.: Imagenet: A
  large-scale hierarchical image database. 2009 IEEE Conference on Computer
  Vision and Pattern Recognition pp. 248--255 (2009)

\bibitem{dosovitskiy2020image}
Dosovitskiy, A., Beyer, L., Kolesnikov, A., Weissenborn, D., Zhai, X.,
  Unterthiner, T., Dehghani, M., Minderer, M., Heigold, G., Gelly, S., et~al.:
  An image is worth 16x16 words: Transformers for image recognition at scale.
  arXiv preprint arXiv:2010.11929  (2020)

\bibitem{Esser_2021_CVPR}
Esser, P., Rombach, R., Ommer, B.: Taming transformers for high-resolution
  image synthesis. In: Proceedings of the IEEE/CVF Conference on Computer
  Vision and Pattern Recognition (CVPR). pp. 12873--12883 (June 2021)

\bibitem{Fergus}
Fergus, R., Fei-Fei, L., Perona, P., Zisserman, A.: Learning object categories
  from internet image searches. Proceedings of the IEEE  \textbf{98}(8),
  1453--1466 (2010). \doi{10.1109/JPROC.2010.2048990}

\bibitem{gal2021stylegannada}
Gal, R., Patashnik, O., Maron, H., Chechik, G., Cohen-Or, D.: Stylegan-nada:
  Clip-guided domain adaptation of image generators (2021)

\bibitem{he2016deep}
He, K., Zhang, X., Ren, S., Sun, J.: Deep residual learning for image
  recognition. In: Proceedings of the IEEE Conference on Computer Vision and
  Pattern Recognition. pp. 770--778 (2016)

\bibitem{Hendrycks_2021_ICCV}
Hendrycks, D., Basart, S., Mu, N., Kadavath, S., Wang, F., Dorundo, E., Desai,
  R., Zhu, T., Parajuli, S., Guo, M., Song, D., Steinhardt, J., Gilmer, J.: The
  many faces of robustness: A critical analysis of out-of-distribution
  generalization. In: Proceedings of the IEEE/CVF International Conference on
  Computer Vision (ICCV). pp. 8340--8349 (October 2021)

\bibitem{Hendrycks_2021_CVPR}
Hendrycks, D., Zhao, K., Basart, S., Steinhardt, J., Song, D.: Natural
  adversarial examples. In: Proceedings of the IEEE/CVF Conference on Computer
  Vision and Pattern Recognition (CVPR). pp. 15262--15271 (June 2021)

\bibitem{NIPS2017_8a1d6947}
Heusel, M., Ramsauer, H., Unterthiner, T., Nessler, B., Hochreiter, S.: Gans
  trained by a two time-scale update rule converge to a local nash equilibrium.
  In: Guyon, I., Luxburg, U.V., Bengio, S., Wallach, H., Fergus, R.,
  Vishwanathan, S., Garnett, R. (eds.) Advances in Neural Information
  Processing Systems. vol.~30. Curran Associates, Inc. (2017),
  \url{https://proceedings.neurips.cc/paper/2017/file/8a1d694707eb0fefe65871369074926d-Paper.pdf}

\bibitem{karras2020analyzing}
Karras, T., Laine, S., Aittala, M., Hellsten, J., Lehtinen, J., Aila, T.:
  Analyzing and improving the image quality of stylegan. In: CVPR. pp.
  8110--8119 (2020)

\bibitem{kim2021diffusionclip}
Kim, G., Ye, J.C.: Diffusionclip: Text-guided image manipulation using
  diffusion models (2021)

\bibitem{Li_2017_ICCV}
Li, A., Jabri, A., Joulin, A., van~der Maaten, L.: Learning visual n-grams from
  web data. In: Proceedings of the IEEE International Conference on Computer
  Vision (ICCV) (Oct 2017)

\bibitem{ty2014coco}
Lin, T.Y., Maire, M., Belongie, S., Hays, J., Perona, P., Ramanan, D.,
  Doll{\'a}r, P., Zitnick, C.L.: Microsoft coco: Common objects in context. In:
  European conference on computer vision. pp. 740--755. Springer (2014)

\bibitem{Liu2021FuseDreamTT}
Liu, X., Gong, C., Wu, L., Zhang, S., Su, H., Liu, Q.: Fusedream: Training-free
  text-to-image generation with improved clip+gan space optimization. ArXiv
  \textbf{abs/2112.01573} (2021)

\bibitem{lundberg2017unified}
Lundberg, S.M., Lee, S.I.: A unified approach to interpreting model
  predictions. In: Advances in Neural Information Processing Systems. pp.
  4765--4774 (2017)

\bibitem{Michel2021Text2MeshTN}
Michel, O.J., Bar-On, R., Liu, R., Benaim, S., Hanocka, R.: Text2mesh:
  Text-driven neural stylization for meshes. ArXiv  \textbf{abs/2112.03221}
  (2021)

\bibitem{Avrahami2021blendeddiffusion}
Omri~Avrahami, D.L., Friedn, O.: Blended diffusion for text-driven editing of
  natural images. arXiv preprint arxiv:2111.14818  (2021)

\bibitem{otsu1979threshold}
Otsu, N.: A threshold selection method from gray-level histograms. IEEE
  transactions on systems, man, and cybernetics  \textbf{9}(1),  62--66 (1979)

\bibitem{patashnik2021styleclip}
Patashnik, O., Wu, Z., Shechtman, E., Cohen-Or, D., Lischinski, D.: Styleclip:
  Text-driven manipulation of stylegan imagery. In: Proceedings of the IEEE/CVF
  International Conference on Computer Vision. pp. 2085--2094 (2021)

\bibitem{radford2021learning}
Radford, A., Kim, J.W., Hallacy, C., Ramesh, A., Goh, G., Agarwal, S., Sastry,
  G., Askell, A., Mishkin, P., Clark, J., et~al.: Learning transferable visual
  models from natural language supervision. arXiv preprint arXiv:2103.00020
  (2021)

\bibitem{Radford2019LanguageMA}
Radford, A., Wu, J., Child, R., Luan, D., Amodei, D., Sutskever, I., et~al.:
  Language models are unsupervised multitask learners. OpenAI blog
  \textbf{1}(8), ~9 (2019)

\bibitem{pmlr-v97-recht19a}
Recht, B., Roelofs, R., Schmidt, L., Shankar, V.: Do {I}mage{N}et classifiers
  generalize to {I}mage{N}et? In: Chaudhuri, K., Salakhutdinov, R. (eds.)
  Proceedings of the 36th International Conference on Machine Learning.
  Proceedings of Machine Learning Research, vol.~97, pp. 5389--5400. PMLR
  (09--15 Jun 2019), \url{https://proceedings.mlr.press/v97/recht19a.html}

\bibitem{Rubinstein}
Rubinstein, M., Joulin, A., Kopf, J., Liu, C.: Unsupervised joint object
  discovery and segmentation in internet images. In: 2013 IEEE Conference on
  Computer Vision and Pattern Recognition. pp. 1939--1946 (2013).
  \doi{10.1109/CVPR.2013.253}

\bibitem{selvaraju2017grad}
Selvaraju, R.R., Cogswell, M., Das, A., Vedantam, R., Parikh, D., Batra, D.:
  Grad-cam: Visual explanations from deep networks via gradient-based
  localization. In: Proceedings of the IEEE international conference on
  computer vision. pp. 618--626 (2017)

\bibitem{shrikumar2016not}
Shrikumar, A., Greenside, P., Shcherbina, A., Kundaje, A.: Not just a black
  box: Learning important features through propagating activation differences.
  arXiv preprint arXiv:1605.01713  (2016)

\bibitem{Tang}
Tang, K., Joulin, A., Li, L.J., Fei-Fei, L.: Co-localization in real-world
  images. In: 2014 IEEE Conference on Computer Vision and Pattern Recognition.
  pp. 1464--1471 (2014). \doi{10.1109/CVPR.2014.190}

\bibitem{Tewel}
Tewel, Y., Shalev, Y., Schwartz, I., Wolf, L.: Zero-shot image-to-text
  generation for visual-semantic arithmetic. CoRR  \textbf{abs/2111.14447}
  (2021), \url{https://arxiv.org/abs/2111.14447}

\bibitem{vaswani2017attention}
Vaswani, A., Shazeer, N., Parmar, N., Uszkoreit, J., Jones, L., Gomez, A.N.,
  Kaiser, {\L}., Polosukhin, I.: Attention is all you need. In: Advances in
  neural information processing systems. pp. 5998--6008 (2017)

\bibitem{Vijayanarasimhan}
Vijayanarasimhan, S., Grauman, K.: Keywords to visual categories:
  Multiple-instance learning forweakly supervised object categorization. In:
  2008 IEEE Conference on Computer Vision and Pattern Recognition. pp.~1--8
  (2008). \doi{10.1109/CVPR.2008.4587632}

\bibitem{NEURIPS2019_3eefceb8}
Wang, H., Ge, S., Lipton, Z., Xing, E.P.: Learning robust global
  representations by penalizing local predictive power. In: Wallach, H.,
  Larochelle, H., Beygelzimer, A., d\textquotesingle Alch\'{e}-Buc, F., Fox,
  E., Garnett, R. (eds.) Advances in Neural Information Processing Systems.
  vol.~32. Curran Associates, Inc. (2019),
  \url{https://proceedings.neurips.cc/paper/2019/file/3eefceb8087e964f89c2d59e8a249915-Paper.pdf}

\bibitem{Wang}
Wang, X.J., Zhang, L., Li, X., Ma, W.Y.: Annotating images by mining image
  search results. IEEE Transactions on Pattern Analysis and Machine
  Intelligence  \textbf{30}(11),  1919--1932 (2008).
  \doi{10.1109/TPAMI.2008.127}

\bibitem{Zabari2021SemanticSI}
Zabari, N., Hoshen, Y.: Semantic segmentation in-the-wild without seeing any
  segmentation examples. ArXiv  \textbf{abs/2112.03185} (2021)

\bibitem{Zhao2020DifferentiableAF}
Zhao, S., Liu, Z., Lin, J., Zhu, J.Y., Han, S.: Differentiable augmentation for
  data-efficient gan training. ArXiv  \textbf{abs/2006.10738} (2020)

\bibitem{zhou2021coop}
Zhou, K., Yang, J., Loy, C.C., Liu, Z.: Learning to prompt for vision-language
  models. arXiv preprint arXiv:2109.01134  (2021)

\bibitem{Zhu2021MindTG}
Zhu, P., Abdal, R., Femiani, J.C., Wonka, P.: Mind the gap: Domain gap control
  for single shot domain adaptation for generative adversarial networks. ArXiv
  \textbf{abs/2110.08398} (2021)

\end{thebibliography}
\appendix

\section{Explainability Method}
\label{app:method}
We create relevance maps for the text tokens (denoted by $\mathbf{R}^{tt}$), and for the image tokens (denoted by $\mathbf{R}^{ii}$) following the method presented in~\cite{Chefer_2021_ICCV}. We initialize the maps as follows:
\begin{align}
    \label{eq:initialize-self}
    \mathbf{R}^{ii} = \mathbb{I}^{i\times i},~~~& \mathbf{R}^{tt} = \mathbb{I}^{t\times t}
\end{align}

Next, we update the relevance maps by a forward pass on the attention layers. We use gradients in order to average across the attention heads, as done in~\cite{Chefer_2021_ICCV}:
\begin{align}
    \label{eq:modified_att}
    \mathbf{\bar{A}} &= \mathbb{E}_h ((\nabla \mathbf{A} \odot \mathbf{A})^+)
\end{align}
where $\odot$ is the Hadamard product, $\nabla \mathbf{A} := \frac{\partial s_{t,i}}{\partial \mathbf{A}}$ for $s_{t,i}$ which is the the similarity score computed by CLIP for the text prompt $t$ with the image $i$, and $\mathbb{E}_h$ is the mean across the heads dimension.
Note that the propagation of gradients by the similarity score allows us to obtain explainability scores for the text that are specific to the input image, i.e. different images induce different explainability scores for each textual token and vice versa.

Finally, to incorporate each layer's explainability map to the accumulated relevancy maps, we use the propagation rule presented in~\cite{Chefer_2021_ICCV} for self-attention layers:
\begin{align}
    \label{eq:self-attention-ii}
    \mathbf{R}^{ii} &\leftarrow  \mathbf{R}^{ii} + \mathbf{\bar{A_i}} \cdot  \mathbf{R}^{ii} \\
    \label{eq:self-attention-tt}
    \mathbf{R}^{tt} &\leftarrow  \mathbf{R}^{tt} + \mathbf{\bar{A_t}} \cdot  \mathbf{R}^{tt}
\end{align}
where $\mathbf{\bar{A_i}}$, $\mathbf{\bar{A_t}}$ are attention relevance maps for image self-attention layers and text attention layers, respectively, which were calculated using Eq.~\ref{eq:modified_att}.

To obtain relevance per each text token, we observe that CLIP uses the $eot$ token as a classification token, thus we simply use the row of $\mathbf{R}^{tt}$ that corresponds to the $eot$ token.

\section{Additional prompt engineering results}
\label{app:prompt-eng}

We present additional results for few-shot classification, which were left out of the main text for brevity. We do not present results for the class-specific configuration due to the required computational resources.

\begin{table}[t]
    \centering
    \begin{tabular}{@{}l@{~}l@{~}c@{~}c@{~}c@{~}c@{~}c@{~}c@{~}c@{~}c@{~}c@{}}
        \toprule
        \begin{small}{Dataset}\end{small} &\multicolumn{2}{c}{ResNet-50} & \multicolumn{2}{c}{ResNet-101} & \multicolumn{2}{c}{ViT-B/16}& \multicolumn{2}{c}{ViT-B/32}  \\
        & CoOp  & Ours &CoOp  & Ours &CoOp  & Ours &CoOp  & Ours \\
        \midrule
        {ImageNet}  & 53.41 & \textbf{57.78} & 56.87& \textbf{61.12}& 63.63 & \textbf{66.19} & 58.23 & \textbf{61.38} \\
        \midrule
        {ImageNetV2} & 46.65 & \textbf{51.20} & 50.17 & \textbf{54.54} & 56.69 & \textbf{58.40} & 50.93 & \textbf{53.82}\\
        \midrule
        {ImageNet-Sketch} & 27.90 & \textbf{32.41}& 34.43& \textbf{37.58}& 41.70 & \textbf{44.66} & 35.33 & \textbf{38.75} \\
        \midrule
        {ImageNet-A} & 20.45 & \textbf{21.89}& 26.63& \textbf{28.74}& 45.58 & \textbf{46.03} & 27.92 & \textbf{30.6} \\
        \midrule
        {ImageNet-R} & 50.31 & \textbf{57.16}& 58.93& \textbf{63.95}& 71.24 & \textbf{73.77} & 61.18 & \textbf{65.49} \\
        \bottomrule
    \end{tabular}
    \smallskip
    \smallskip
    \caption{1-shot accuracy (in percentage) of CLIP~\cite{radford2021learning} with prompts produced by the method of \cite{zhou2021coop} (CoOp) and by our explainability-guided variant, with class name tokens positioned at the end of the prompts. Results are averaged over 3 random seeds.}\label{table:promptengineering_end}
\end{table}

\subsection{Prompt engineering with class name at the end of prompt}
\label{app:prompt-eng-end}

As mentioned in Sec.~\ref{sec:prompt_exp} in the main text, Zhou et al.~\cite{zhou2021coop} proposed two options for positioning the class name tokens in the optimized prompts, which achieved similar results. The first has the class name positioned in the middle of the prompt, i.e.: t = v1,...,v8,{label},v9,...,v16, where v1,...,v16 are
the prompt tokens, and the second has the class name located at the end, i.e.: t = v1,...,v16,{label}.
The results of our explainability-based method for the middle positioning variant are reported in the main text. The results for the end positioning variant are presented in Tab.~\ref{table:promptengineering_end}. As can be seen, our method consistently outperforms the original method of \cite{zhou2021coop} with the end positioning {configuration, similar} to the middle positioning one.

\subsection{Few-shot prompt engineering}
\label{app:prompt-eng-few-shot}
Although this work focuses on the 1-shot scenario for prompt engineering, we report the results for 2-shot and 4-shot optimization as well. Tab.~\ref{table:promptengineering_2shots} and \ref{table:promptengineering_4shots} present the accuracy achieved with the method proposed in \cite{zhou2021coop} and with our method for the unified prompt configuration, where a single prompt is optimized for all class names.
As can be seen in Tab.~\ref{table:promptengineering_2shots}, our method consistently outperforms the original method of \cite{zhou2021coop} with a significant margin across the different datasets and backbones in the 2-shot scenario.
A similar situation occurs in the 4-shot scenario, as can be seen in Tab.~\ref{table:promptengineering_4shots}. 
Note that for some of the visual backbones, the accuracy our method achieves for 1-shot optimization surpasses the accuracy achieved by the original method in the 2-shot scenario, and the results for our method in the 2-shot scenario are very close to the results of the original method in the 4-shot scenario. Since each $i$-shot, for $i \in \{2,4\}$ duplicates the size of the training set compared to $(i-1)$-shot, this is a strong indication for the effectiveness of our method in improving generalization.

\begin{table}[t]
    \centering
    \begin{tabular}{@{}l@{~}l@{~}c@{~}c@{~}c@{~}c@{~}c@{~}c@{~}c@{~}c@{~}c@{}}
        \toprule
        \begin{small}{Dataset}\end{small} &\multicolumn{2}{c}{ResNet-50} & \multicolumn{2}{c}{ResNet-101} & \multicolumn{2}{c}{ViT-B/16}& \multicolumn{2}{c}{ViT-B/32}  \\
        & CoOp  & Ours &CoOp  & Ours &CoOp  & Ours &CoOp  & Ours \\
        \midrule
        {ImageNet}  & 56.70 & \textbf{58.42} & 60.59& \textbf{62.71}& 65.94 & \textbf{67.72} & 60.59 & \textbf{62.71} \\
        \midrule
        {ImageNetV2} & 49.99 & \textbf{51.99} & 52.85 & \textbf{54.95} & 58.48 & \textbf{60.87} & 52.66 & \textbf{55.10}\\
        \midrule
        {ImageNet-Sketch} & 29.31 & \textbf{31.46}& 36.03& \textbf{37.95}& 43.06 & \textbf{45.79} & 37.18 & \textbf{39.70} \\
        \midrule
        {ImageNet-A} & 21.85 & \textbf{22.19}& 26.98& \textbf{28.83}& 45.69 & \textbf{46.96} & 28.68 & \textbf{30.87}\\
        \midrule
        {ImageNet-R} & 52.03 & \textbf{56.34}& 60.73& \textbf{63.54}& 71.15 & \textbf{74.22} & 61.84 & \textbf{64.92} \\
        \bottomrule
    \end{tabular}
    \smallskip
    \smallskip
    \caption{2-shot accuracy (in percentage) of CLIP~\cite{radford2021learning} with prompts produced by the method \cite{zhou2021coop} (CoOp) and by our explainability-guided variant. Results are averaged over 3 random seeds.}
    \label{table:promptengineering_2shots}
\end{table}

\begin{table}[t]
    \centering
    \begin{tabular}{@{}l@{~}l@{~}c@{~}c@{~}c@{~}c@{~}c@{~}c@{~}c@{~}c@{~}c@{}}
        \toprule
        \begin{small}{Dataset}\end{small} &\multicolumn{2}{c}{ResNet-50} & \multicolumn{2}{c}{ResNet-101} & \multicolumn{2}{c}{ViT-B/16}& \multicolumn{2}{c}{ViT-B/32}  \\
        & CoOp  & Ours &CoOp  & Ours &CoOp  & Ours &CoOp  & Ours \\
        \midrule
        {ImageNet}  & 59.50 & \textbf{59.87} & 62.84& \textbf{63.59}& 68.20 & \textbf{68.90} & 63.12 & \textbf{64.14} \\
        \midrule
        {ImageNetV2} & 52.04 & \textbf{52.22} & 55.31 & \textbf{56.25} & 61.39 & \textbf{61.85} & 54.72 & \textbf{56.57}\\
        \midrule
        {ImageNet-Sketch} & 30.89 & \textbf{32.45}& 36.97& \textbf{38.84}& 44.40 & \textbf{45.92} & 38.31 & \textbf{40.18} \\
        \midrule
        {ImageNet-A} & 21.83 & \textbf{22.48}& 28.37& \textbf{29.93}& 46.93 & \textbf{48.23} & 29.50 & \textbf{31.14} \\
        \midrule
        {ImageNet-R} & 54.03 & \textbf{55.82}& 60.61& \textbf{63.78}& 71.36 & \textbf{74.13} & 62.47 & \textbf{65.68} \\
        \bottomrule
    \end{tabular}
    \smallskip
    \smallskip
    \caption{4-shot accuracy (in percentage) of CLIP~\cite{radford2021learning} with prompts produced by the method of \cite{zhou2021coop} (CoOp) and by our explainability-guided variant. Results are averaged over 3 random seeds.}
    \label{table:promptengineering_4shots}
\end{table}

\begin{figure}[t!]
\centering
\includegraphics[width=1\linewidth, clip]
{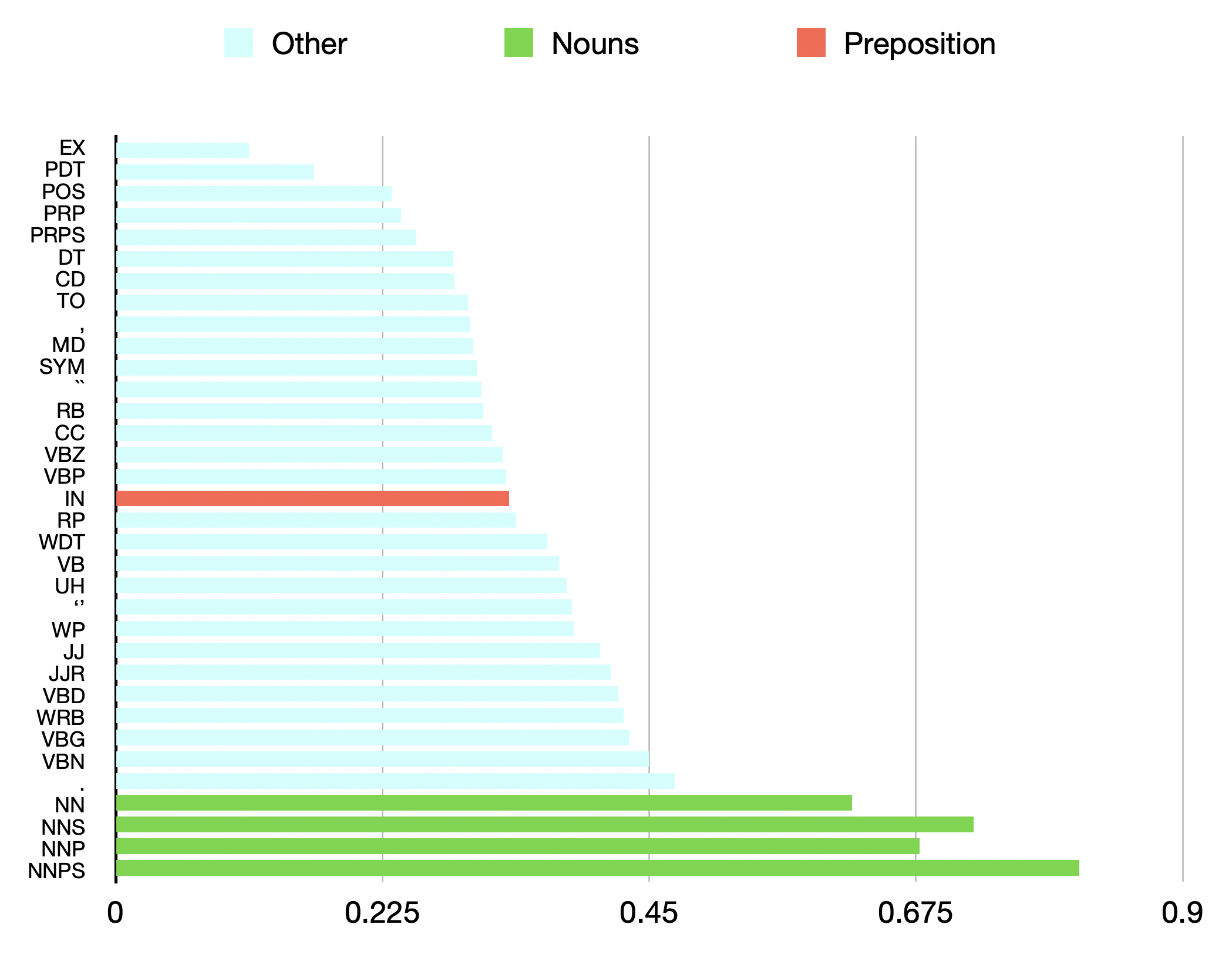}\\
\caption{
{ Average textual explainability score for different speech parts (POS), calculated over the matching pairs of text and image in MSCOCO~\cite{ty2014coco} validation set. The POS that describe entities are colored in green, and preposition is colored in red. As can be seen, when predicting that truly matching image and caption are similar, the nouns (objects) are most relevant to the prediction, significantly more that the spatial positioning.} \label{fig:L2I_POS_dist}}
\end{figure}

\section{Relevance scores distribution over different POS}
\label{app:pos}

In order to understand the limitations of CLIP-guided optimization in general, we study the importance of different speech parts (POS) to the prediction delivered by CLIP. We calculate the textual explainability scores for all matching pairs of image and caption in MSCOCO~\cite{ty2014coco} validation set. The explainability scores for each caption are divided by the maximal explainability score in it, to allow comparison of the relevancy of words between different sentences. We extract the POS of each caption using the part-of-speech tagging architecture of \cite{akbik2018coling} with the Flair framework~\cite{akbik2019flair}, and average the relevance scores of each POS over the entire MSCOCO~\cite{ty2014coco} validation set. Fig.~\ref{fig:L2I_POS_dist} shows the average relevance score of different speech parts, for all speech parts that appear at least 20 times in the data.
As can be seen in Fig.~\ref{fig:L2I_POS_dist} CLIP bases its similarity scores on the nouns significantly more than any other speech part in the text, including prepositions (IN) and adjectives (JJ), meaning that the existence of a given object in both the caption and the image is more important for the similarity score prediction than its {detailed} attributes or spatial position. This observation explains why CLIP-guided optimizations often fail to follow spatial positions described in the input text prompt.

\section{Text-guided image generation}
\label{app:fusedream}

\begin{figure*}[t!]
~~~~~~Selected basis by \cite{Liu2021FuseDreamTT} {~~~~~~~~~~~~}  \cite{Liu2021FuseDreamTT} {~~~~~~~} Selected basis by our method {~}  Our result\\

\centering
\includegraphics[width=1\linewidth, clip]
{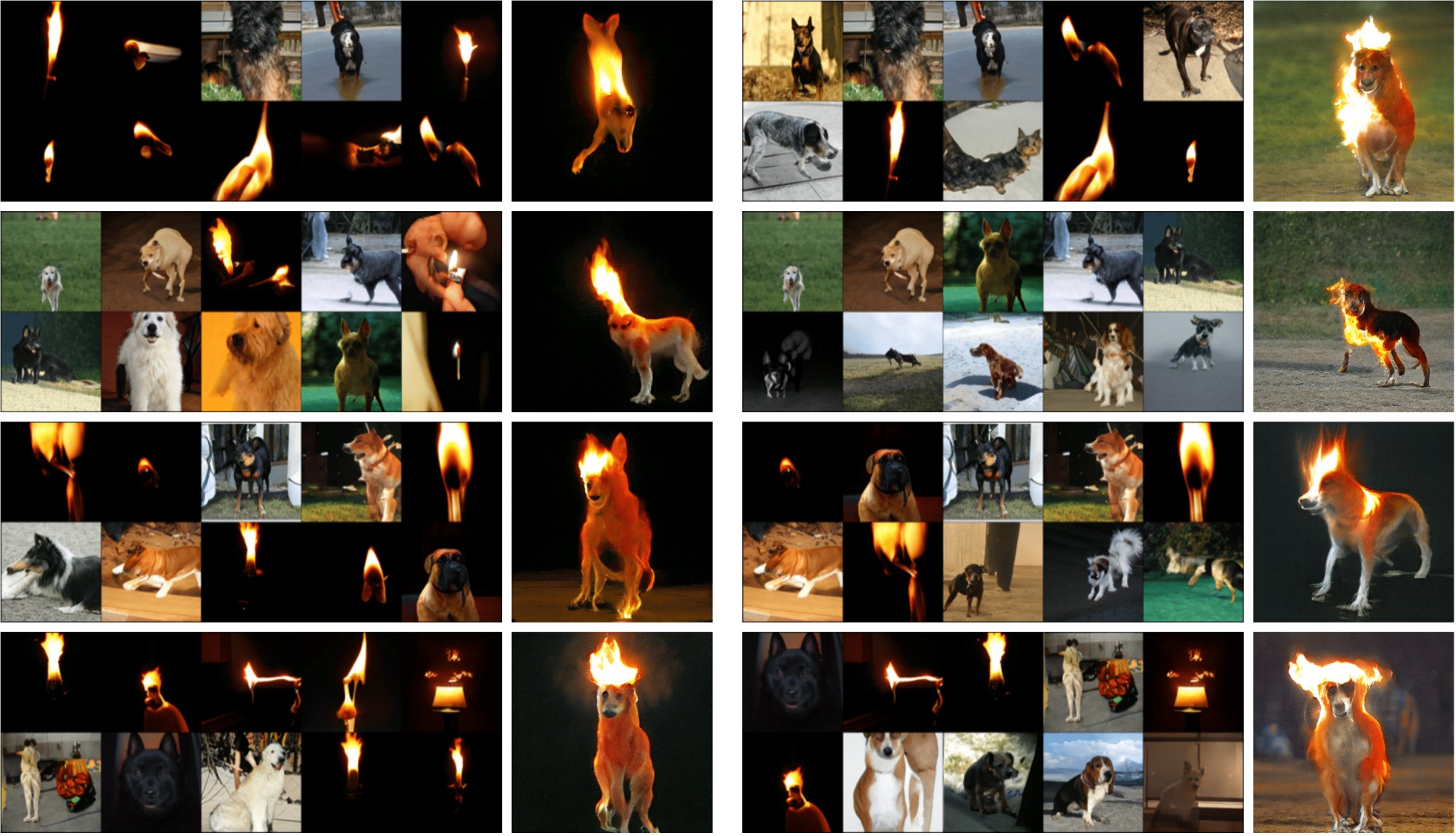}\\

\caption{Generated images conditioned on the prompt ``a photo of a flaming dog" using FuseDream~\cite{Liu2021FuseDreamTT} with and without our method. The word ``flaming" is very dominant and the original FuseDream basis selection results with mostly images of flames. This unbalanced basis limits the ability of the optimization process to generate the dog and leads to a small variation both in the background and the dog itself. \label{fig:a_flaming_dog}}
\end{figure*}

\begin{figure*}[t!]
  \centering
  \includegraphics[width=0.95\linewidth, clip]
  {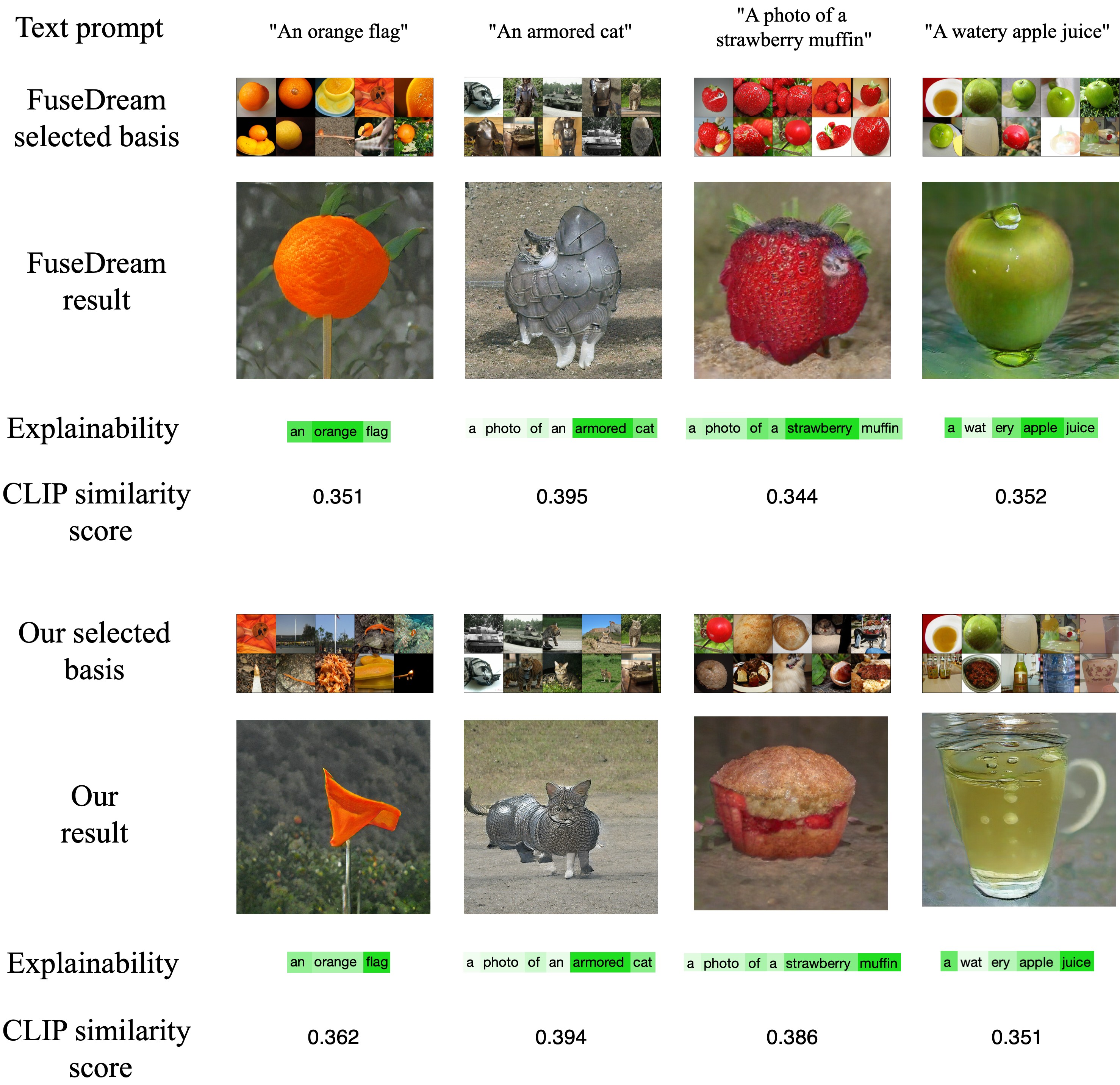} \\
\caption{Examples of selected bases and resulting generated images with FuseDream~\cite{Liu2021FuseDreamTT} with and without our explainability guidance, on prompts with compound nouns presented in the user study. The basis selection of \cite{Liu2021FuseDreamTT} results in basis images that severely neglect relevant nouns from the prompts such as flag, cat, juice, and muffin, as reflected in both the images and their corresponding explainability scores. In contrast, our method selects basis vectors that better correspond to the input prompts, resulting in generated images that follow the semantic meaning of the texts.}
\label{fig:fd_our_prompts}
\end{figure*}

We describe a method for text-guided image generation that is similar in spirit to the methods presented in the main text. For brevity, and since the benefit of the proposed improvement is most apparent in the specific case of compound nouns, we exclude the results from the main text. Moreover, unlike the loss-based approach used in the paper, the intervention here is only in the retrieval mechanism used for the basis selection stage of the generative scheme.

The FuseDream method~\cite{Liu2021FuseDreamTT} improves upon a direct application of BigGAN~\cite{brock2018large} for CLIP-guided text-based image generation. FuseDream employs a modified CLIP score based on image augmentations, as follows:
\begin{align}
\label{eq:augclip}
      \text{AUGCLIP}(t,i) = \mathbb{E}_{i'\sim  \pi(\cdot | i)}[\text{CLIP}(t,i')],
\end{align}
where $i'$ is a random augmentation of $i$ drawn from the distribution $\pi(\cdot | i)$. The data augmentations are adopted from \cite{Zhao2020DifferentiableAF}.

{Since the optimization process is essentially a traversal in the latent space of a pretrained BigGAN, aimed to locate the latent vector from which BigGAN generates an image that matches the text prompt, a key contribution of \cite{Liu2021FuseDreamTT} regards the initialization of this traversal. Instead of initializing the optimization process with a randomly sampled latent vector, the} 
FuseDream method randomly generates $M$ images using BigGAN, and selects the vectors in the BigGAN latent space that generate the $k$ images with the highest AUGCLIP score as a basis $B$.  {This basis selection phase allows for shorter and easier navigation of the latent space, as it is initialized closer to the target vector to be reached.} Let $B=\{v_1,...,v_k\}$ be the selected basis (where $v_1,...,v_k$ are latent vectors). The FuseDream method optimizes the following:
\begin{align}
\label{eq:augclip2}
      \max_{\{\epsilon_i, w_i\}_{i=1}^k} \text{AUGCLIP}\left(t, g\left(\sum_{j=1}^k w_i \epsilon_i\right)\right),
\end{align}
where $g$ is the BigGAN generator, the vectors are initialized to be the basis vectors: $\epsilon_j = v_j$, and the coefficients are initialized as: $w_j = \frac{1}{k}$. The final output image after the optimization is: $i_{out} = g\left(\sum_{j=1}^k w_i \epsilon_i\right)$.
{Since the basis selection relies solely on the similarity score predicted by CLIP, it can exhibit a neglecting behavior, resulting in a retrieved basis that is unbalanced and fails to represent the semantic meaning of the input text. An example can be seen in}
Fig.~\ref{fig:fd_our_prompts}, for the phrase ``a photo of a strawberry muffin'', all images chosen as the basis of FuseDream feature strawberries, {which leads to a sub-optimal initialization of the optimization process, resulting in} output images that are shaped like a strawberry, rather than a muffin. This can be attributed to the fact that for the images corresponding to strawberries, CLIP predicts a similarity score that is based mostly on the word ``strawberry'', disregarding the word ``muffin''. Fig.~\ref{fig:a_flaming_dog} presents another example, for the basis produced for the prompt ``a photo of a flaming dog'', in which the word ``flaming'' is emphasized over ``dog''. 
In order to overcome this sensitivity, our method accounts for the explainability scores of the produced similarity, instead of simply considering the pure similarity scores, by adding $\mathcal{L}_{expl}$ from Eq.~\ref{eq:expl_gen_loss} in the main text to the AUGCLIP score to select the basis vectors. This ensures that the basis vectors indeed reflect the {entire semantic content} of the textual prompt.

\textbf{Choosing the set of semantic words $S$:}
Eq.~\ref{eq:expl_gen_loss} uses a set of semantic words $S$ to ensure that the similarity score predicted by CLIP is based on the true semantic meaning of the input text. In order to produce the set $S$ automatically, our method uses a part-of-speech tagging architecture~\cite{akbik2018coling} with the Flair framework~\cite{akbik2019flair} to automatically extract all words that correspond to nouns in the input text prompt $t$. Next, the FuseDream~\cite{Liu2021FuseDreamTT} method is used to generate an image $i$ by the description $t$, and a relevance score $\mathcal{R}_{expl}$ is computed for each word in $t$ w.r.t. the generated image $i$, as described in Sec.~\ref{sec:explainability} in the main text. The set $S$ is defined as follows:
\begin{align}
\label{eq:choose_S}
      S=\{t_i \in t_{noun} \text{ s.t. } \mathcal{R}_{expl}(t_i) < 0.7\},
\end{align}
where $t_{noun}$ is the set of words in $t$ that are classified by the part-of-speech tagger as nouns. Therefore, the loss in Eq.~\ref{eq:expl_gen_loss} emphasizes all nouns in the input prompt with a relevance score lower than 0.7 for the produced image $i$. Intuitively, each noun represents an object that should appear in the image $i$, therefore nouns that have a low relevance score either do not appear or appear only partially in the image $i$. The loss in Eq.~\ref{eq:expl_gen_loss} ensures that the neglected objects of $i$ will appear once it is applied to the modified basis selection. In our experiments, we set: $\lambda_{expl}$=0.1.

\subsection{FuseDream experiments}
We conduct all our experiments with the default setting from the FuseDream~\cite{Liu2021FuseDreamTT} code base, using a BigGAN~\cite{brock2018large} generator for $512$ resolution images, with $10$ basis vectors for each image generation.
To evaluate the visual quality of both methods, we conduct a user study, with $46$ participants. The study presents the users with $53$ textual prompts, for which we generate corresponding images with FuseDream~\cite{Liu2021FuseDreamTT}, and with our modified method described above. Users are asked to choose the image that corresponds best to the textual description or mark both as equally successful. $46$ of the presented prompts are the visual examples presented in \cite{Liu2021FuseDreamTT}, so as to compare our method against the most visually pleasing results by FuseDream. We focus on the examples where our method produces a different result than FuseDream{, meaning the set $S$ contains at least one word}. The other $7$ prompts are prompts containing compound nouns, such as "strawberry muffin" or "orange kimono", {as we found these cases tend to exhibit neglect by CLIP.}
For each example, we use the majority of the answers to determine which method produced the best image. In $33$ of the $53$ images ($62.26\%$) the participants ruled that our method produced an image that corresponds better to the input text. Additionally, the average ratings for the $7$ prompts that contain compound nouns show that in $86.6\%$  of the cases, users voted for our method as producing the results most compatible with the textual descriptions. See Fig.~\ref{fig:fd_our_prompts} for a full comparison between the selected basis and the final images produced by our method and FuseDream for $4$ of the $7$ prompts containing compound nouns. As can be seen, our method selects basis images that correspond better to the target semantic prompt; therefore it generates images that reflect the textual descriptions.   

Next, following the metrics presented in \cite{Liu2021FuseDreamTT}, we present the Fr\'echet inception distance (FID)~\cite{NIPS2017_8a1d6947} on a subset of $30,000$ randomly sampled prompts from the MSCOCO~\cite{ty2014coco} validation set. As can be seen in Tab~\ref{tab:fid}, our method slightly improves the FID score over FuseDream, indicating that, in accordance with the user study, our method either preserves the successful results of FuseDream, or improves the produced images, in cases {where the similarity-based optimization fails to capture the entire textual description}

\begin{table}[t]
    \smallskip
    \centering
   \begin{tabular}{@{}l@{~}l@{~}c@{~}c@{~}c@{~}c@{}}
        \toprule
        Basis size& \multicolumn{1}{c}{FuseDream} & \multicolumn{1}{c}{Ours}  \\
        \midrule
        M=5 & 21.26 & \textbf{20.66} \\
        M=10 & 24.67 & \textbf{23.96}\\
        \bottomrule
    \end{tabular}
    \smallskip
    \smallskip
\caption{{FID~\cite{NIPS2017_8a1d6947} (lower is better) calculated on images generated with FuseDream~\cite{Liu2021FuseDreamTT} and with our method according to $30,000$ randomly sampled prompts from the MSCOCO validation set. Both FuseDream and our method use a pretrained BigGAN~\cite{brock2018large} generator for $512$ resolution images. Our method slightly improves the results of FuseDream.}}
    \label{tab:fid}
\end{table}

\section{StyleCLIP user study results}
\label{app:sc}
\begin{figure*}[t!]
  \centering
  \begin{tabular}{c@{~}c@{~}c@{~}}
  Original Image & StyleCLIP & Ours \\

\includegraphics[width=0.15\linewidth, clip]{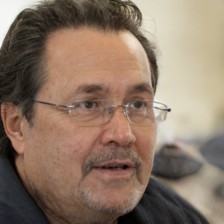}&
\includegraphics[width=0.15\linewidth, clip]{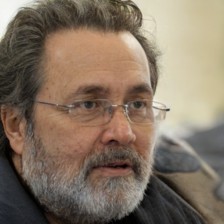}&
\includegraphics[width=0.15\linewidth, clip]{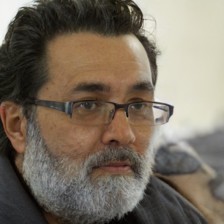} \\

\includegraphics[width=0.15\linewidth, clip]{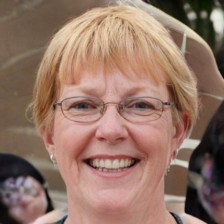}&
\includegraphics[width=0.15\linewidth, clip]{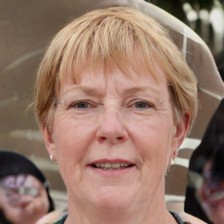}&
\includegraphics[width=0.15\linewidth, clip]{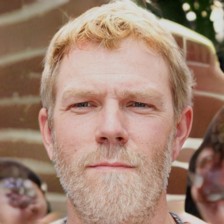} \\

\includegraphics[width=0.15\linewidth, clip]{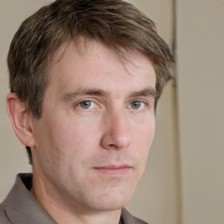}&
\includegraphics[width=0.15\linewidth, clip]{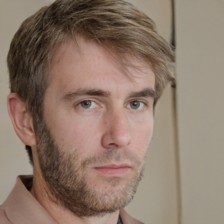}&
\includegraphics[width=0.15\linewidth, clip]{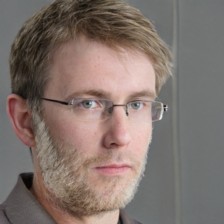} \\

\includegraphics[width=0.15\linewidth, clip]{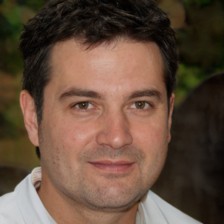}&
\includegraphics[width=0.15\linewidth, clip]{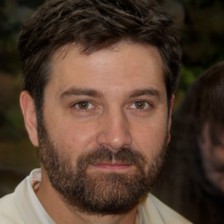}&
\includegraphics[width=0.15\linewidth, clip]{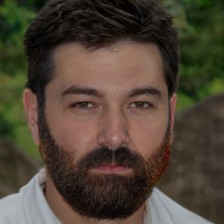}\\

\includegraphics[width=0.15\linewidth, clip]{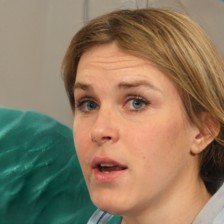}&
\includegraphics[width=0.15\linewidth, clip]{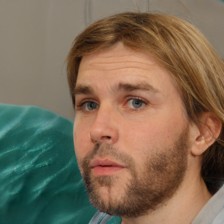}&
\includegraphics[width=0.15\linewidth, clip]{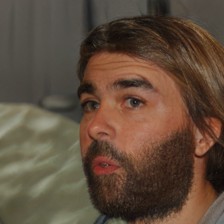}\\

\includegraphics[width=0.15\linewidth, clip]{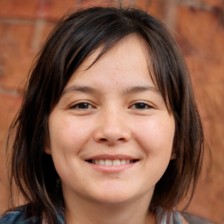}&
\includegraphics[width=0.15\linewidth, clip]{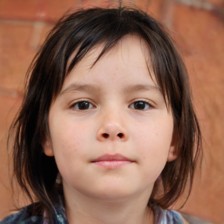}&
\includegraphics[width=0.15\linewidth, clip]{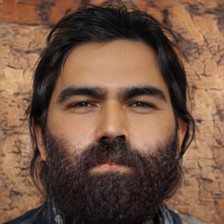}\\

\includegraphics[width=0.15\linewidth, clip]{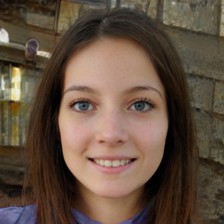}&
\includegraphics[width=0.15\linewidth, clip]{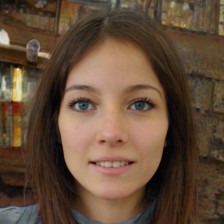}&
\includegraphics[width=0.15\linewidth, clip]{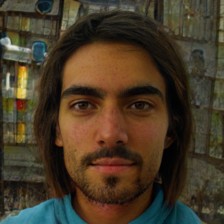}

\end{tabular}
\caption{Results of the StyleCLIP~\cite{patashnik2021styleclip} user study on the prompt ``A man with a beard." The examples above represent the seeds where our method generates a different result than that of StyleCLIP, out of the $20$ random seeds selected for the study. For the other seeds, our method produces the same result as StyleCLIP ($\lambda_{expl}=0$). \label{fig:styleclip-beard}}
\end{figure*}

\begin{figure*}[t!]
  \centering
  \begin{tabular}{c@{~}c@{~}c@{~}}
  Original Image & StyleCLIP & Ours \\

\includegraphics[width=0.15\linewidth, clip]{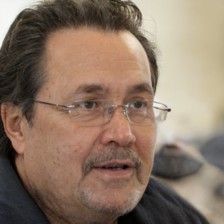}&
\includegraphics[width=0.15\linewidth, clip]{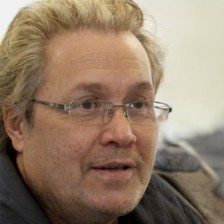}&
\includegraphics[width=0.15\linewidth, clip]{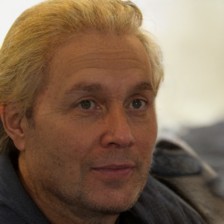} \\

\includegraphics[width=0.15\linewidth, clip]{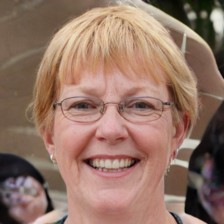}&
\includegraphics[width=0.15\linewidth, clip]{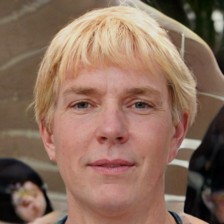}&
\includegraphics[width=0.15\linewidth, clip]{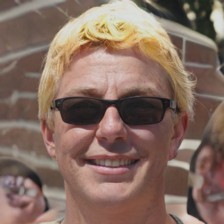} \\

\includegraphics[width=0.15\linewidth, clip]{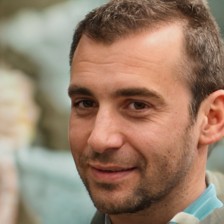}&
\includegraphics[width=0.15\linewidth, clip]{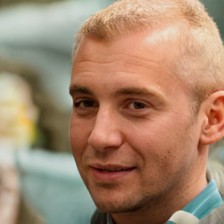}&
\includegraphics[width=0.15\linewidth, clip]{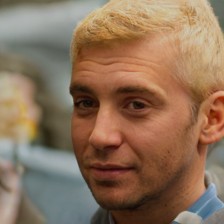} \\

\includegraphics[width=0.15\linewidth, clip]{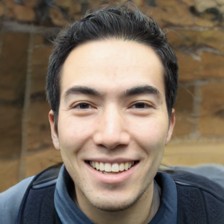}&
\includegraphics[width=0.15\linewidth, clip]{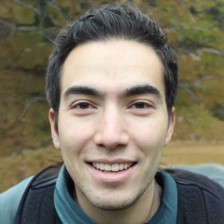}&
\includegraphics[width=0.15\linewidth, clip]{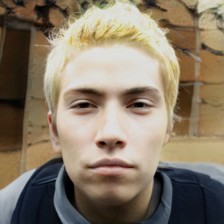}\\

\includegraphics[width=0.15\linewidth, clip]{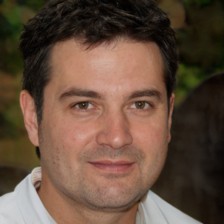}&
\includegraphics[width=0.15\linewidth, clip]{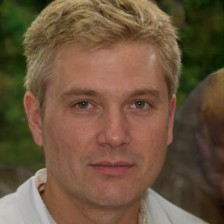}&
\includegraphics[width=0.15\linewidth, clip]{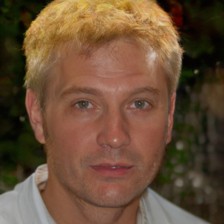}\\

\includegraphics[width=0.15\linewidth, clip]{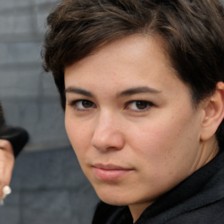}&
\includegraphics[width=0.15\linewidth, clip]{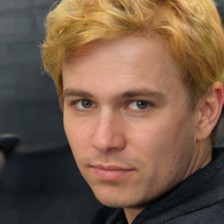}&
\includegraphics[width=0.15\linewidth, clip]{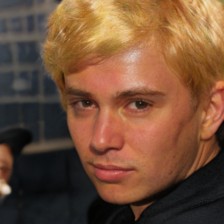}\\

\includegraphics[width=0.15\linewidth, clip]{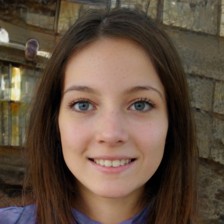}&
\includegraphics[width=0.15\linewidth, clip]{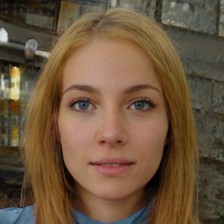}&
\includegraphics[width=0.15\linewidth, clip]{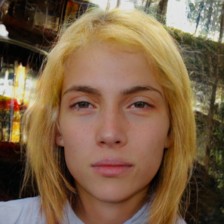}

\end{tabular}
\caption{Results of the StyleCLIP~\cite{patashnik2021styleclip} user study on the prompt ``A blond man." The examples above represent the seeds where our method generates a different result than that of StyleCLIP, out of the $20$ random seeds selected for the study. For the other seeds, our method produces the same result as StyleCLIP ($\lambda_{expl}=0$). \label{fig:styleclip-blond}}
\end{figure*}

\begin{figure*}[t!]
  \centering
  \begin{tabular}{c@{~}c@{~}c@{~}}
  Original Image & StyleCLIP & Ours \\

\includegraphics[width=0.2\linewidth, clip]{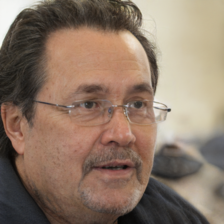}&
\includegraphics[width=0.2\linewidth, clip]{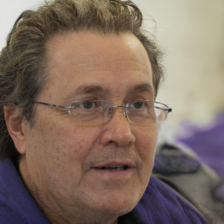}&
\includegraphics[width=0.2\linewidth, clip]{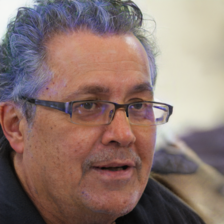} \\

\includegraphics[width=0.2\linewidth, clip]{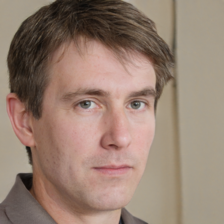}&
\includegraphics[width=0.2\linewidth, clip]{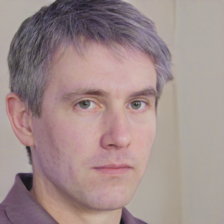}&
\includegraphics[width=0.2\linewidth, clip]{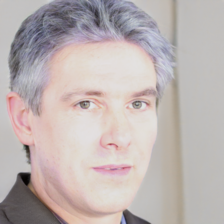} \\

\includegraphics[width=0.2\linewidth, clip]{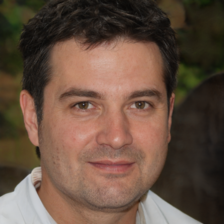}&
\includegraphics[width=0.2\linewidth, clip]{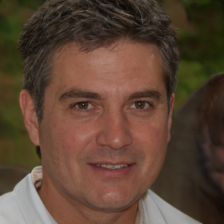}&
\includegraphics[width=0.2\linewidth, clip]{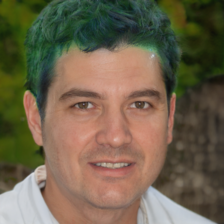}\\

\includegraphics[width=0.2\linewidth, clip]{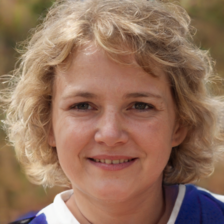}&
\includegraphics[width=0.2\linewidth, clip]{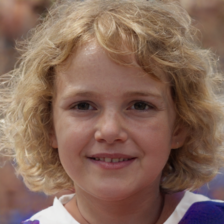}&
\includegraphics[width=0.2\linewidth, clip]{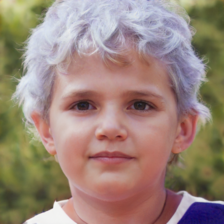}\

\end{tabular}
\caption{Results of the StyleCLIP~\cite{patashnik2021styleclip} user study on the prompt ``A person with purple hair." The examples above represent the seeds where our method generates a different result than that of StyleCLIP, out of the $20$ random seeds selected for the study. For the other seeds, our method produces the same result as StyleCLIP ($\lambda_{expl}=0$). \label{fig:styleclip-purple}}
\end{figure*}

\begin{figure*}[t!]
  \centering
  \begin{tabular}{c@{~}c@{~}c@{~}}
  Original Image & StyleCLIP & Ours \\
\includegraphics[width=0.2\linewidth, clip]{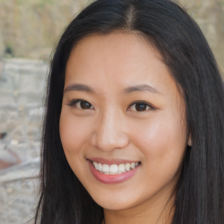}&
\includegraphics[width=0.2\linewidth, clip]{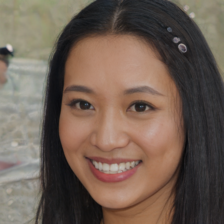}&
\includegraphics[width=0.2\linewidth, clip]{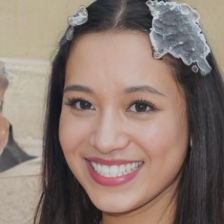} \\

\includegraphics[width=0.2\linewidth, clip]{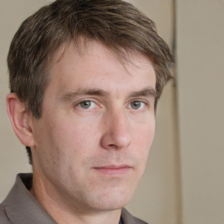}&
\includegraphics[width=0.2\linewidth, clip]{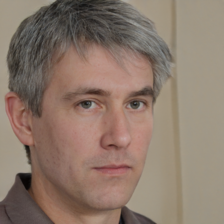}&
\includegraphics[width=0.2\linewidth, clip]{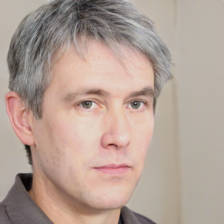} \\

\includegraphics[width=0.2\linewidth, clip]{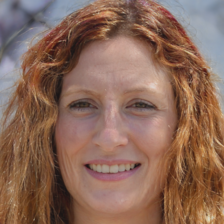}&
\includegraphics[width=0.2\linewidth, clip]{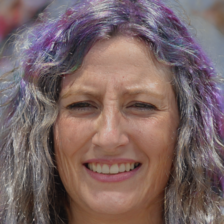}&
\includegraphics[width=0.2\linewidth, clip]{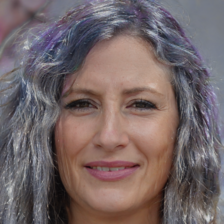}\\

\includegraphics[width=0.2\linewidth, clip]{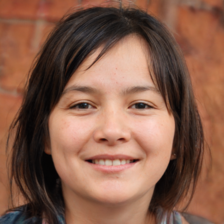}&
\includegraphics[width=0.2\linewidth, clip]{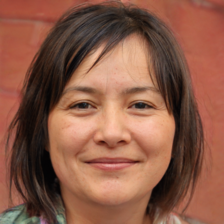}&
\includegraphics[width=0.2\linewidth, clip]{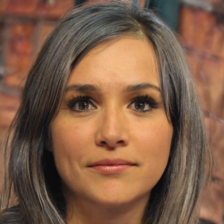}\

\end{tabular}
\caption{Results of the StyleCLIP~\cite{patashnik2021styleclip} user study on the prompt ``A person with grey hair." The examples above represent the seeds where our method generates a different result than that of StyleCLIP, out of the $20$ random seeds selected for the study. For the other seeds, our method produces the same result as StyleCLIP ($\lambda_{expl}=0$). \label{fig:styleclip-grey}}
\end{figure*}

Tab.~\ref{fig:userstudy_SD} presents the StyleCLIP user study results with standard deviation per metric. Notice that standard deviations tend to be high since some of the manipulations performed by both methods fail, resulting in low quality scores. As can be seen, our standard deviation for the quality score is consistently lower than StyleCLIP's (with the exception of prompt (c) where both are very similar), indicating that our manipulations are more stable.  Fig.~\ref{fig:styleclip-beard},~\ref{fig:styleclip-blond},~\ref{fig:styleclip-purple},~\ref{fig:styleclip-grey} present the seeds from the StyleCLIP user study where our method generates a different result than that of StyleCLIP, out of the $20$ random seeds used for the study. For the other seeds, our method produces the same result as StyleCLIP ($\lambda_{expl}=0$). As mentioned in the main text, for prompts that entail a change of identity such as "a blond \emph{man}" and "a \emph{man} with a beard", our method causes a more significant identity change in accordance with the prompt.

\begin{table*}[t!]  
\vskip 0.15in
\centering
\begin{tabular}{ ccccc }

\begin{tabular*}{0.45\linewidth}{@{\extracolsep{\fill}}lcc}
        \toprule
        Method&Quality&Identity \\
        \midrule
        SC & 2.92 ($\pm$ 1.86) & \textbf{3.61} ($\pm$ 0.68) \\
        Ours & \textbf{4.28} ($\pm$ 0.54) & 2.23 ($\pm$ 0.72) \\
        \bottomrule
    \end{tabular*} & 
\begin{tabular*}{0.45\linewidth}{@{\extracolsep{\fill}}lcc}
        \toprule
        Method&Quality&Identity \\
        \midrule
        SC & 1.17 ($\pm$ 0.18)& \textbf{4.13} ($\pm$ 0.27) \\
        Ours & \textbf{2.29} ($\pm$ 1.39) & 3.51 ($\pm$ 0.8) \\
        \bottomrule
    \end{tabular*} \\
    (a) & (b) \\
    \\
    \begin{tabular*}{0.45\linewidth}{@{\extracolsep{\fill}}lcc}
        \toprule
        Method&Quality&Identity \\
        \midrule
        SC & 3.93 ($\pm$ 1.40)& \textbf{3.67} ($\pm$ 0.80)\\
        Ours & \textbf{4.28} ($\pm$ 0.50)& 2.63 ($\pm$ 0.86) \\
        \bottomrule
    \end{tabular*} &
    \begin{tabular*}{0.45\linewidth}{@{\extracolsep{\fill}}lcc}
        \toprule
        Method&Quality&Identity \\
        \midrule
        SC & 2.59 ($\pm$ 1.80)& \textbf{3.82} ($\pm$ 0.78) \\
        Ours & \textbf{3.27}($\pm$ 1.51) & 3.10($\pm$ 1.01) \\
        \bottomrule
    \end{tabular*} \\
     (c) & (d)
    \end{tabular}
    \caption{Results of the user study comparing text-based image editing with StyleCLIP (SC) and our method on 4 different textual prompts. (a) ``A man with a beard'', (b) ``A person with purple hair'', (c) ``A blond man'', (d) ``A person with grey hair''. Quality refers to the similarity between the prompt and the manipulation; Identity refers to the identity preservation of the manipulation.  Scores are averaged across 20 random seeds, on a scale of 1-5 (higher is better). Notice that standard deviations tend to be high since some of the manipulations preformed by both methods fail, resulting in low quality scores. }
    \label{fig:userstudy_SD}
\end{table*}

\section{Zero-shot text to image generation with spatial conditioning}
\label{app:l2i}

\subsection{Parameter sensitivity for $\lambda_{expl}$, $T$ and $temp$}
\label{app:l2i_hyperparams}

The sensitivity of our method for spatially conditioned image generation to its  hyperparameters $\lambda_{expl}$, $T$ and $temp$ is studied in Tab.~\ref{tab:l2i_ablation_temp},~\ref{tab:l2i_ablation_T}, and~\ref{tab:l2i_ablation_lambda} respectively.
The temperature  $temp$ is used along with a $sigmoid$ function to transform the continuous and normalized explainabiltiy scores into semi-binarized map. As can be seen in Tab.~\ref{tab:l2i_ablation_temp}, setting $temp=1$ reduces all DETR metrics significantly. However, values in a wide range of 10-40 all produce reasonable results.

{As can bee seen in Tab.~\ref{tab:l2i_ablation_T} and \ref{tab:l2i_ablation_lambda}, our method works well with different values for $\lambda_{expl}$ and $T$, resulting with significantly better metrics than the baselines.
}

\subsection{Zero-shot text to image generation with spatial conditioning visualizations}
\label{app:l2i_examples}
\begin{figure*}[t!]
  \centering
  \begin{tabular}{ccccc}
   Input & Textual &Similarity- & Similarity- & Ours\\
   conditioning & conditioning & based & based 2 &\\
\includegraphics[width=0.18\linewidth,clip] 
{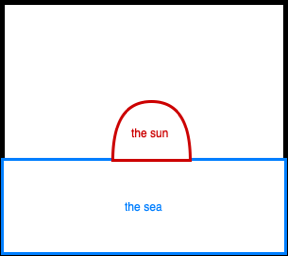}&
\includegraphics[width=0.18\linewidth,clip] 
{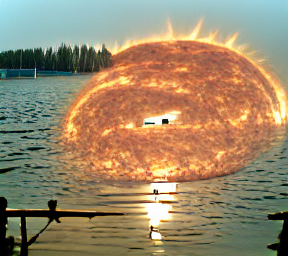} &
\includegraphics[width=0.18\linewidth,clip] 
{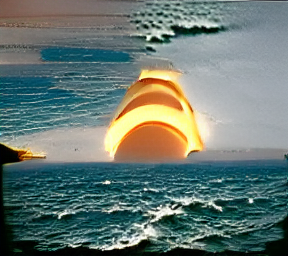} &
\includegraphics[width=0.18\linewidth,clip] 
{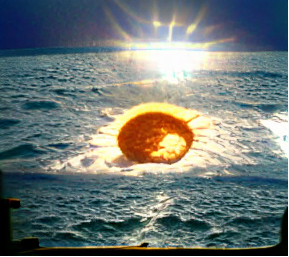}&
\includegraphics[width=0.18\linewidth,clip] 
{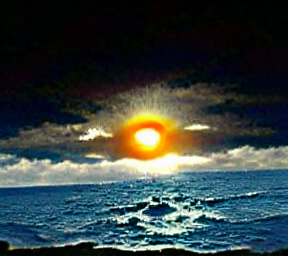}\\

\includegraphics[width=0.18\linewidth,clip]
{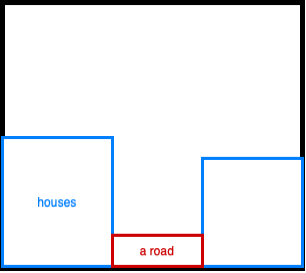}&
\includegraphics[width=0.18\linewidth,clip]
{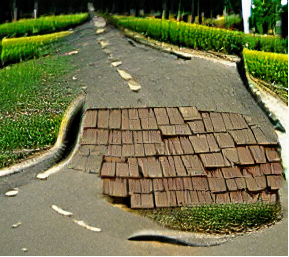}&
\includegraphics[width=0.18\linewidth,clip]
{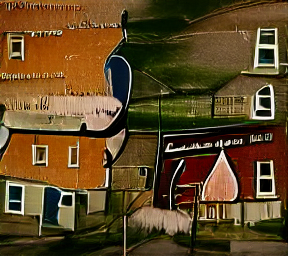}&
\includegraphics[width=0.18\linewidth,clip]
{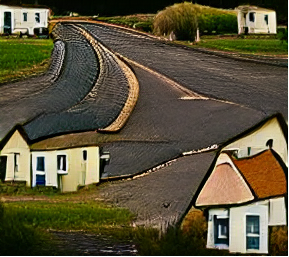}&
\includegraphics[width=0.18\linewidth,clip]
{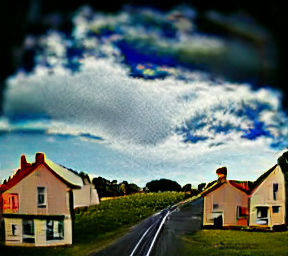}\\

\includegraphics[width=0.18\linewidth,clip]
{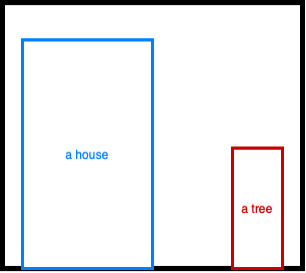}&
\includegraphics[width=0.18\linewidth,clip]
{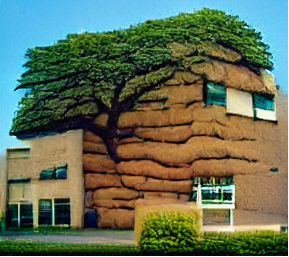}&
\includegraphics[width=0.18\linewidth,clip]
{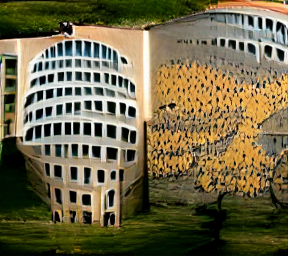}&
\includegraphics[width=0.18\linewidth,clip]
{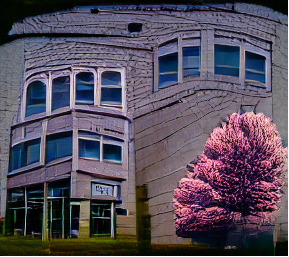}&
\includegraphics[width=0.18\linewidth,clip]
{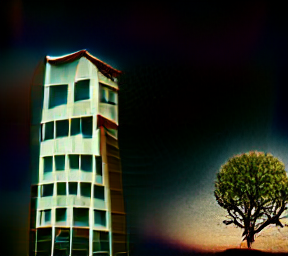}\\

\includegraphics[width=0.18\linewidth,clip]
{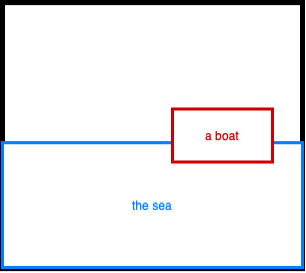}&
\includegraphics[width=0.18\linewidth,clip]
{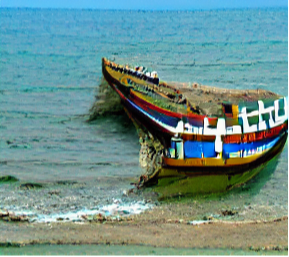}&
\includegraphics[width=0.18\linewidth,clip]
{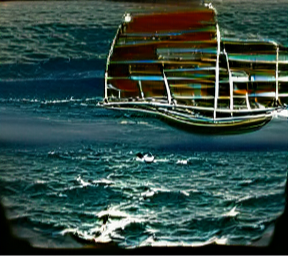}&
\includegraphics[width=0.18\linewidth,clip]
{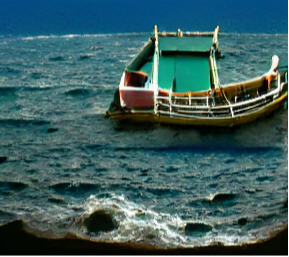}&
\includegraphics[width=0.18\linewidth,clip]
{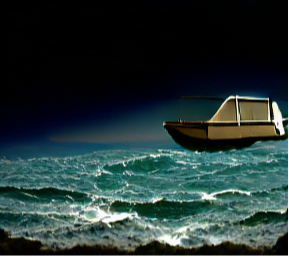}\\

\includegraphics[width=0.18\linewidth,clip]
{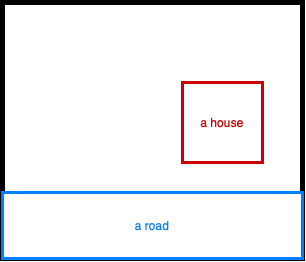}&
\includegraphics[width=0.18\linewidth,clip]
{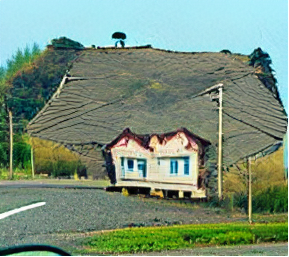}&
\includegraphics[width=0.18\linewidth,clip]
{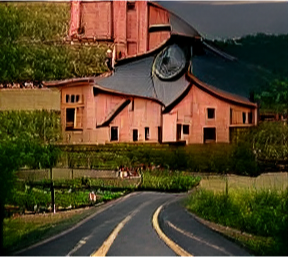}&
\includegraphics[width=0.18\linewidth,clip]
{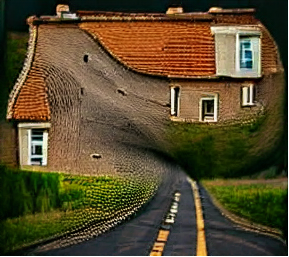}&
\includegraphics[width=0.18\linewidth,clip]
{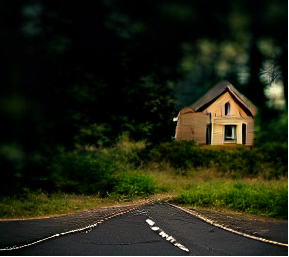}\\

\includegraphics[width=0.18\linewidth,clip]
{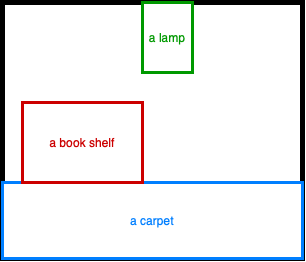}&
\includegraphics[width=0.18\linewidth,clip]
{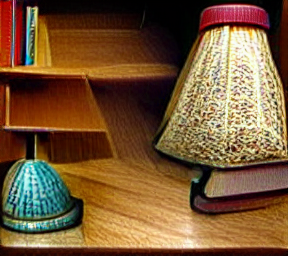}&
\includegraphics[width=0.18\linewidth,clip]
{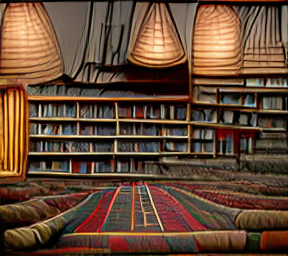}&
\includegraphics[width=0.18\linewidth,clip]
{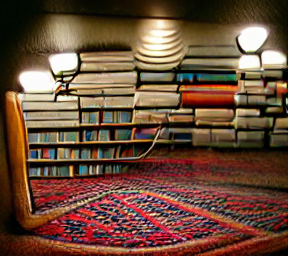}&
\includegraphics[width=0.18\linewidth,clip]
{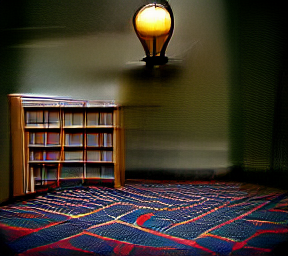}\\

\includegraphics[width=0.18\linewidth,clip]
{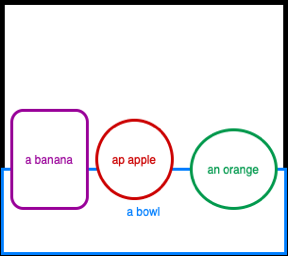}&
\includegraphics[width=0.18\linewidth,clip] 
{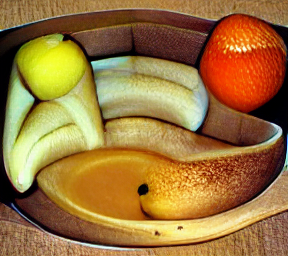} &
\includegraphics[width=0.18\linewidth,clip] 
{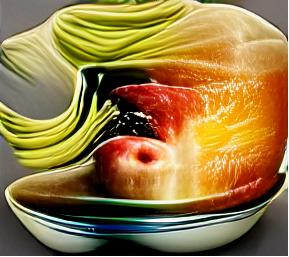} &
\includegraphics[width=0.18\linewidth,clip] 
{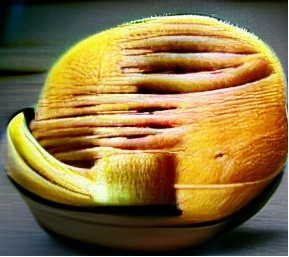}&
\includegraphics[width=0.18\linewidth,clip] 
{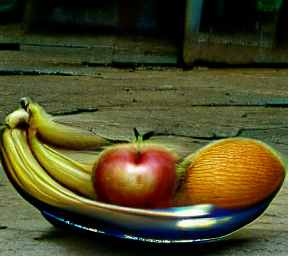}\\
\includegraphics[width=0.18\linewidth,clip] 
{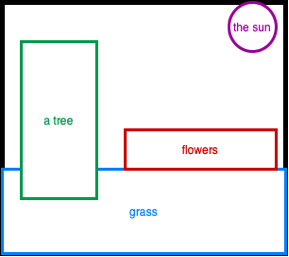}&
\includegraphics[width=0.18\linewidth,clip] 
{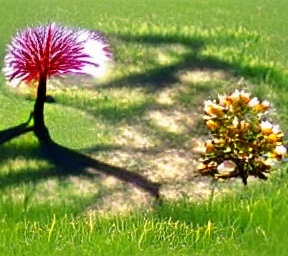} &
\includegraphics[width=0.18\linewidth,clip] 
{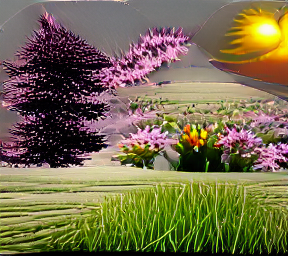} &
\includegraphics[width=0.18\linewidth,clip] 
{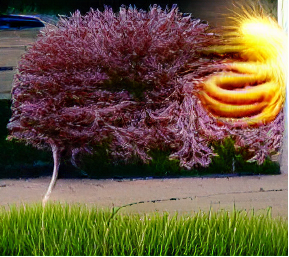}&
\includegraphics[width=0.18\linewidth,clip] 
{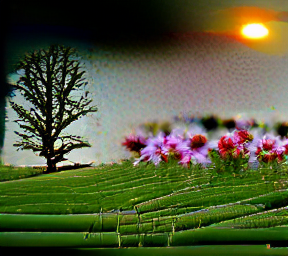}\\

\includegraphics[width=0.18\linewidth,clip]
{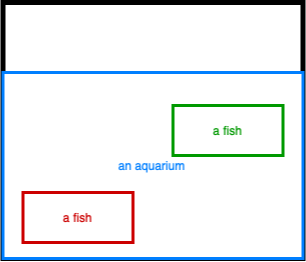}&
\includegraphics[width=0.18\linewidth,clip]
{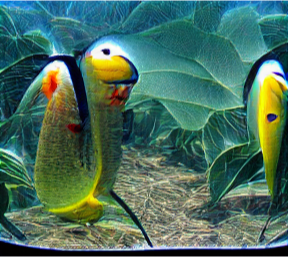}&
\includegraphics[width=0.18\linewidth,clip]
{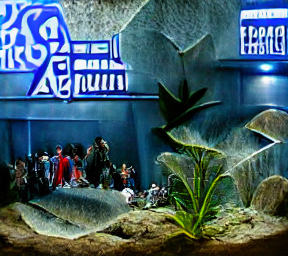}&
\includegraphics[width=0.18\linewidth,clip]
{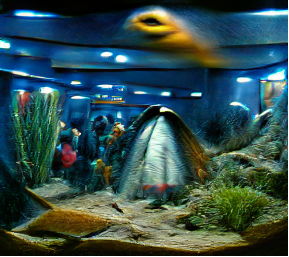}&
\includegraphics[width=0.18\linewidth,clip]
{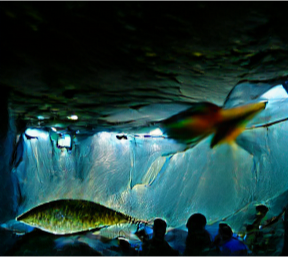}\\

\end{tabular}
\caption{Examples of spatially conditioned images generated with the similarity-based baselines and with our explainability-based method.} \label{fig:l2i_more_examples}
\end{figure*}

\begin{figure*}[t!]
  \centering
  \begin{tabular}{ccccc}
   Input & Textual &Similarity- & Similarity- & Ours\\
   conditioning & conditioning & based & based 2 &\\

\includegraphics[width=0.18\linewidth,clip]
{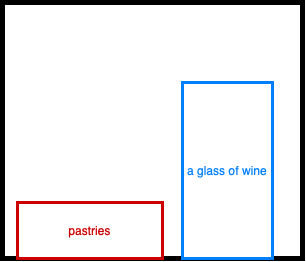}&
\includegraphics[width=0.18\linewidth,clip]
{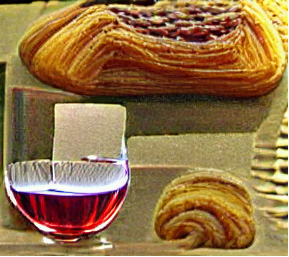}&
\includegraphics[width=0.18\linewidth,clip]
{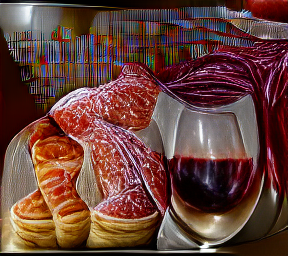}&
\includegraphics[width=0.18\linewidth,clip]
{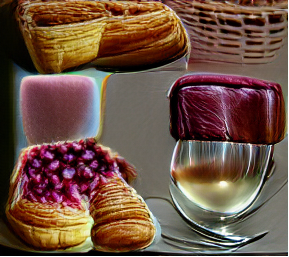}&
\includegraphics[width=0.18\linewidth,clip]
{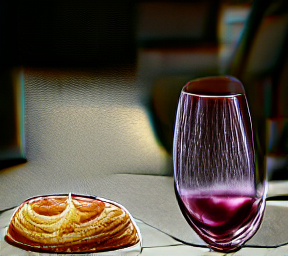}\\

\includegraphics[width=0.18\linewidth,clip]
{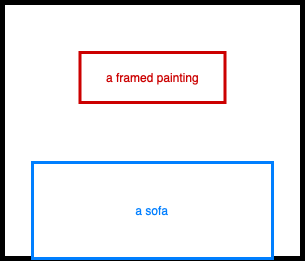}&
\includegraphics[width=0.18\linewidth,clip]
{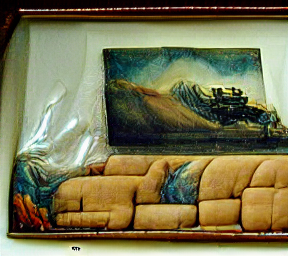}&
\includegraphics[width=0.18\linewidth,clip]
{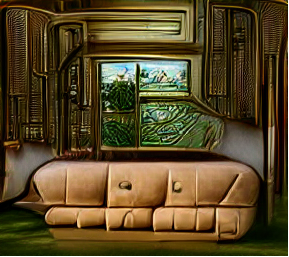}&
\includegraphics[width=0.18\linewidth,clip]
{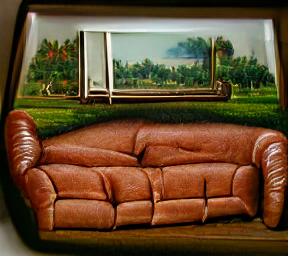}&
\includegraphics[width=0.18\linewidth,clip]
{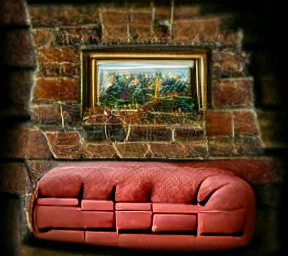}\\

\includegraphics[width=0.18\linewidth,clip]
{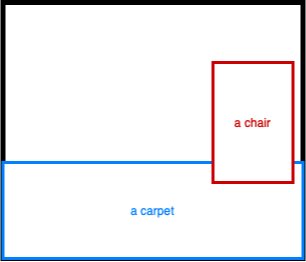}&
\includegraphics[width=0.18\linewidth,clip]
{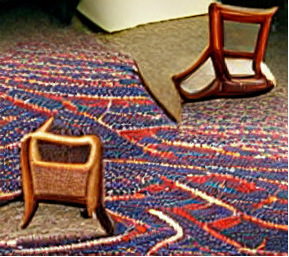}&
\includegraphics[width=0.18\linewidth,clip]
{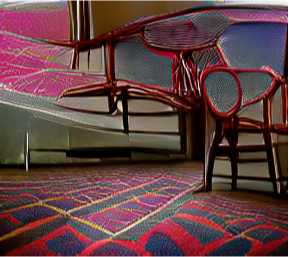}&
\includegraphics[width=0.18\linewidth,clip]
{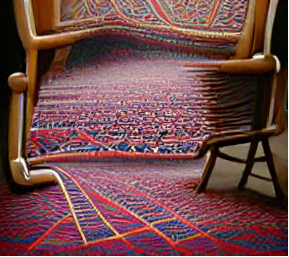}&
\includegraphics[width=0.18\linewidth,clip]
{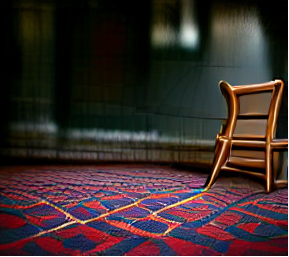}\\

\includegraphics[width=0.18\linewidth,clip]
{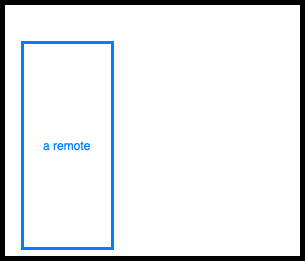}&
\includegraphics[width=0.18\linewidth,clip]
{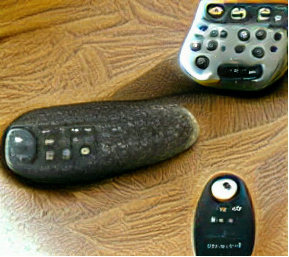}&
\includegraphics[width=0.18\linewidth,clip]
{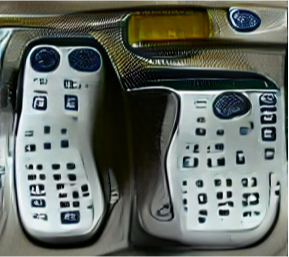}&
\includegraphics[width=0.18\linewidth,clip]
{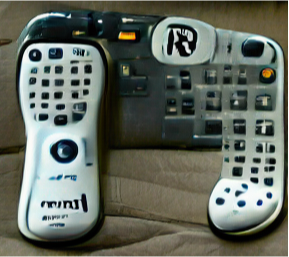}&
\includegraphics[width=0.18\linewidth,clip]
{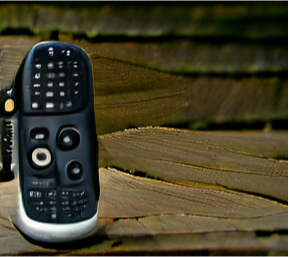}\\

\includegraphics[width=0.18\linewidth,clip]
{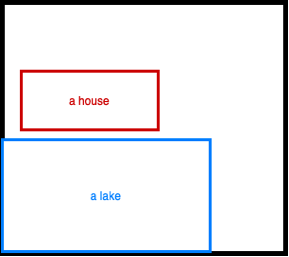}&
\includegraphics[width=0.18\linewidth,clip]
{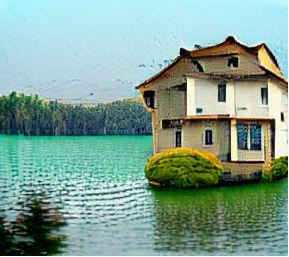}&
\includegraphics[width=0.18\linewidth,clip]
{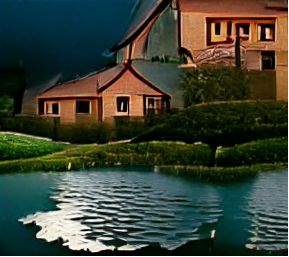}&
\includegraphics[width=0.18\linewidth,clip]
{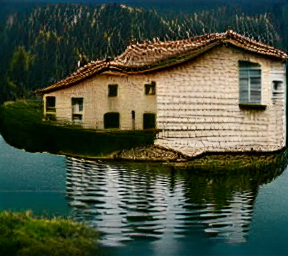}&
\includegraphics[width=0.18\linewidth,clip]
{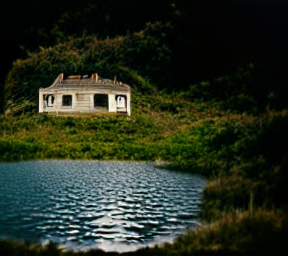}\\

\includegraphics[width=0.18\linewidth,clip]
{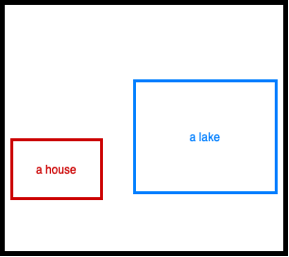}&
\includegraphics[width=0.18\linewidth,clip]
{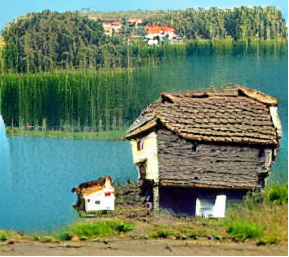}&
\includegraphics[width=0.18\linewidth,clip]
{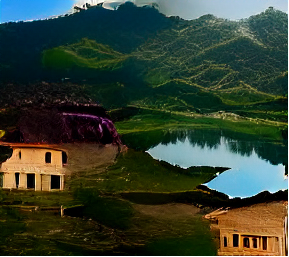}&
\includegraphics[width=0.18\linewidth,clip]
{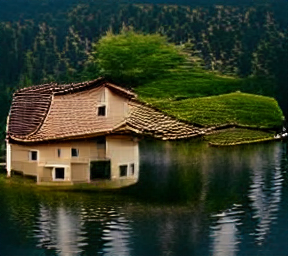}&
\includegraphics[width=0.18\linewidth,clip]
{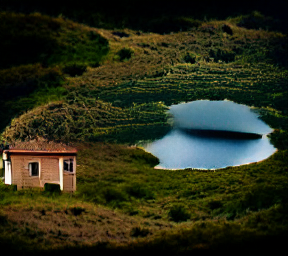}\\

\includegraphics[width=0.18\linewidth,clip]
{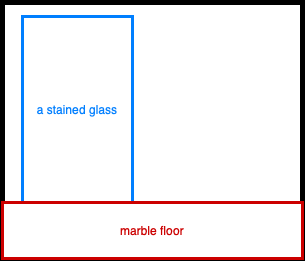}&
\includegraphics[width=0.18\linewidth,clip]
{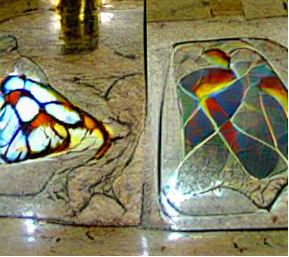}&
\includegraphics[width=0.18\linewidth,clip]
{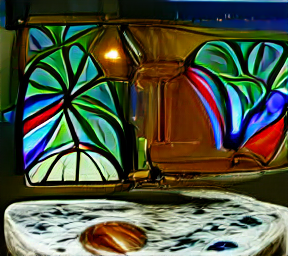}&
\includegraphics[width=0.18\linewidth,clip]
{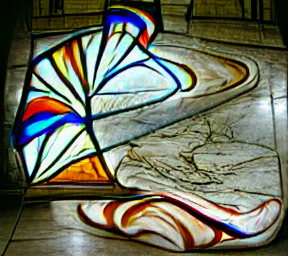}&
\includegraphics[width=0.18\linewidth,clip]
{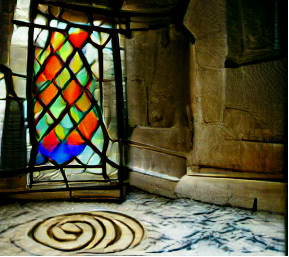}\\

\includegraphics[width=0.18\linewidth,clip]
{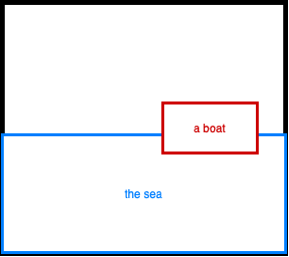}&
\includegraphics[width=0.18\linewidth,clip]
{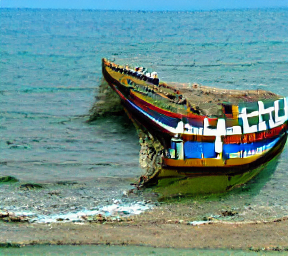}&
\includegraphics[width=0.18\linewidth,clip]
{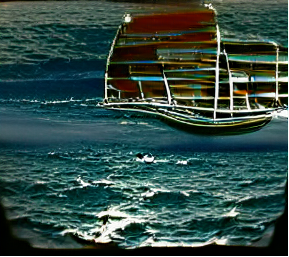}&
\includegraphics[width=0.18\linewidth,clip]
{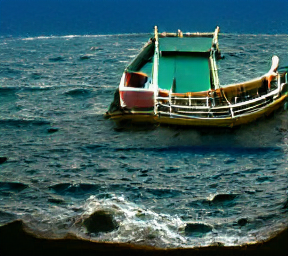}&
\includegraphics[width=0.18\linewidth,clip]
{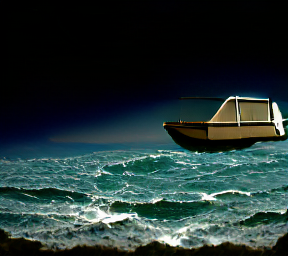}\\

\end{tabular}
\caption{Examples of spatially conditioned images generated with the similarity-based baselines and with our explainability-based method. \label{fig:l2i_more_examples2}}
\end{figure*}

\begin{figure*}[t!]
  \centering
  \begin{tabular}{c@{~~~}c@{~}c@{~}c@{~~~}c@{~}c@{~}c@{~~~}c@{~}c@{~}c}
\multicolumn{1}{c}{spatial} &
\multicolumn{3}{c}{Similarity-} &
\multicolumn{3}{c}{Similarity-} &
\multicolumn{3}{c}{Ours} \\
\multicolumn{1}{c}{conditioning} &
\multicolumn{3}{c}{based 1} &
\multicolumn{3}{c}{based 2} & \\

\includegraphics[width=0.08\linewidth,clip]
{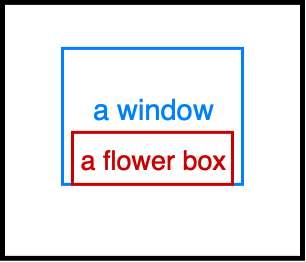}&
\includegraphics[width=0.08\linewidth,clip]
{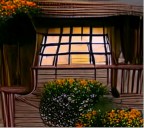}&
\includegraphics[width=0.08\linewidth,clip]
{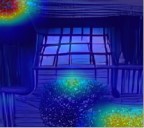}&
\includegraphics[width=0.08\linewidth,clip]
{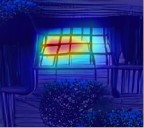}&
\includegraphics[width=0.08\linewidth,clip]
{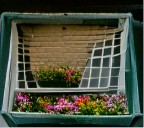}&
\includegraphics[width=0.08\linewidth,clip]
{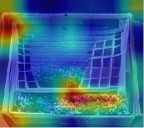}&
\includegraphics[width=0.08\linewidth,clip]
{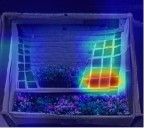}&
\includegraphics[width=0.08\linewidth,clip]
{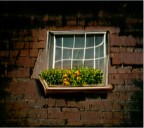}&
\includegraphics[width=0.08\linewidth,clip]
{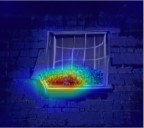}&
\includegraphics[width=0.08\linewidth,clip]
{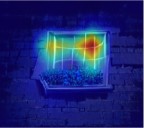}\\

\includegraphics[width=0.08\linewidth,clip]
{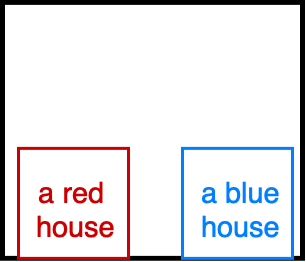}&
\includegraphics[width=0.08\linewidth,clip]
{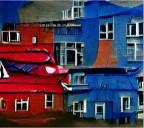}&
\includegraphics[width=0.08\linewidth,clip]
{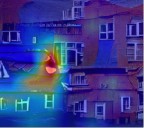}&
\includegraphics[width=0.08\linewidth,clip]
{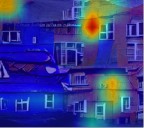}&
\includegraphics[width=0.08\linewidth,clip]
{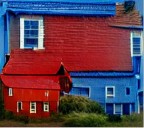}&
\includegraphics[width=0.08\linewidth,clip]
{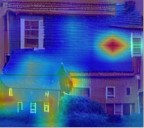}&
\includegraphics[width=0.08\linewidth,clip]
{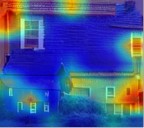}&
\includegraphics[width=0.08\linewidth,clip]
{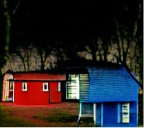}&
\includegraphics[width=0.08\linewidth,clip]
{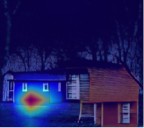}&
\includegraphics[width=0.08\linewidth,clip]
{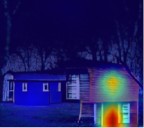}\\

\includegraphics[width=0.08\linewidth,clip]
{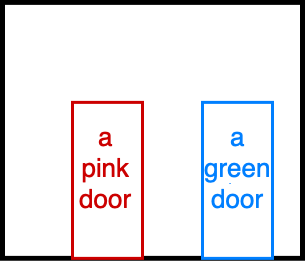}&
\includegraphics[width=0.08\linewidth,clip]
{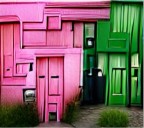}&
\includegraphics[width=0.08\linewidth,clip]
{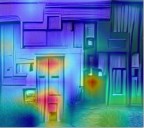}&
\includegraphics[width=0.08\linewidth,clip]
{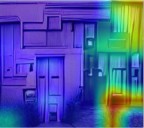}&
\includegraphics[width=0.08\linewidth,clip]
{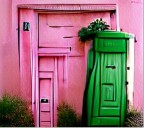}&
\includegraphics[width=0.08\linewidth,clip]
{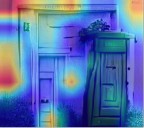}&
\includegraphics[width=0.08\linewidth,clip]
{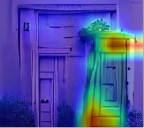}&
\includegraphics[width=0.08\linewidth,clip]
{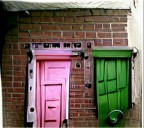}&
\includegraphics[width=0.08\linewidth,clip]
{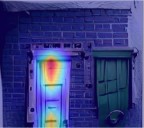}&
\includegraphics[width=0.08\linewidth,clip]
{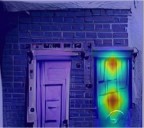}\\

\includegraphics[width=0.08\linewidth,clip]
{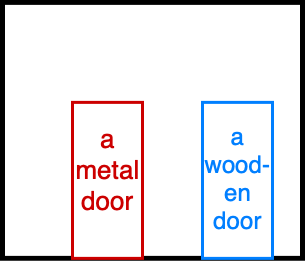}&
\includegraphics[width=0.08\linewidth,clip]
{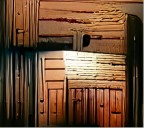}&
\includegraphics[width=0.08\linewidth,clip]
{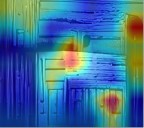}&
\includegraphics[width=0.08\linewidth,clip]
{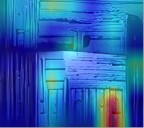}&
\includegraphics[width=0.08\linewidth,clip]
{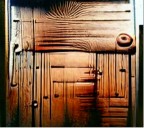}&
\includegraphics[width=0.08\linewidth,clip]
{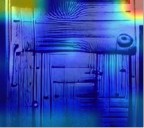}&
\includegraphics[width=0.08\linewidth,clip]
{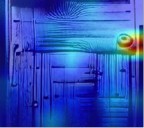}&
\includegraphics[width=0.08\linewidth,clip]
{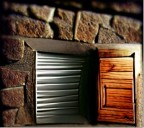}&
\includegraphics[width=0.08\linewidth,clip]
{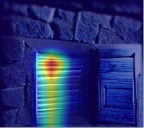}&
\includegraphics[width=0.08\linewidth,clip]
{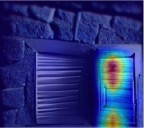}\\

\includegraphics[width=0.08\linewidth,clip]
{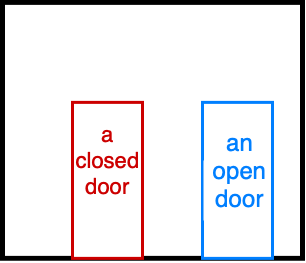}&
\includegraphics[width=0.08\linewidth,clip]
{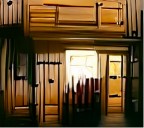}&
\includegraphics[width=0.08\linewidth,clip]
{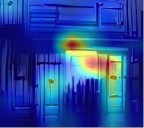}&
\includegraphics[width=0.08\linewidth,clip]
{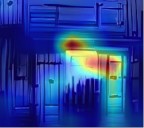}&
\includegraphics[width=0.08\linewidth,clip]
{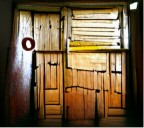}&
\includegraphics[width=0.08\linewidth,clip]
{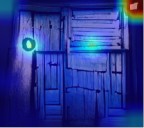}&
\includegraphics[width=0.08\linewidth,clip]
{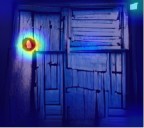}&
\includegraphics[width=0.08\linewidth,clip]
{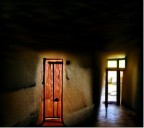}&
\includegraphics[width=0.08\linewidth,clip]
{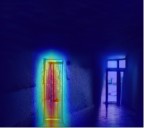}&
\includegraphics[width=0.08\linewidth,clip]
{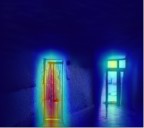}\\

\includegraphics[width=0.08\linewidth,clip]
{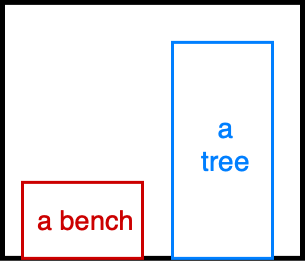}&
\includegraphics[width=0.08\linewidth,clip]
{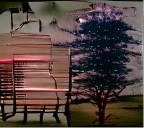}&
\includegraphics[width=0.08\linewidth,clip]
{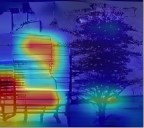}&
\includegraphics[width=0.08\linewidth,clip]
{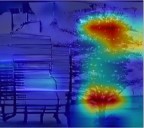}&
\includegraphics[width=0.08\linewidth,clip]
{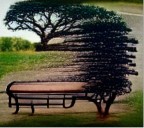}&
\includegraphics[width=0.08\linewidth,clip]
{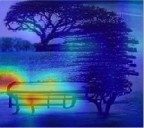}&
\includegraphics[width=0.08\linewidth,clip]
{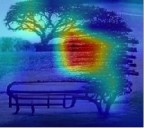}&
\includegraphics[width=0.08\linewidth,clip]
{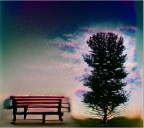}&
\includegraphics[width=0.08\linewidth,clip]
{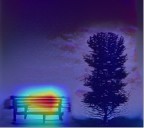}&
\includegraphics[width=0.08\linewidth,clip]
{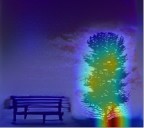}\\

\includegraphics[width=0.08\linewidth,clip]
{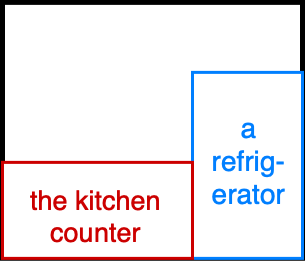}&
\includegraphics[width=0.08\linewidth,clip]
{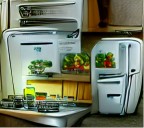}&
\includegraphics[width=0.08\linewidth,clip]
{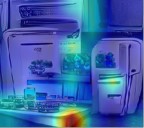}&
\includegraphics[width=0.08\linewidth,clip]
{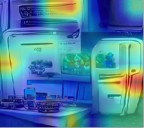}&
\includegraphics[width=0.08\linewidth,clip]
{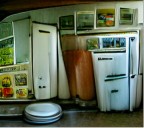}&
\includegraphics[width=0.08\linewidth,clip]
{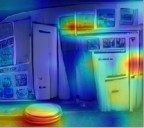}&
\includegraphics[width=0.08\linewidth,clip]
{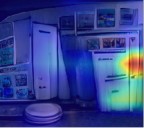}&
\includegraphics[width=0.08\linewidth,clip]
{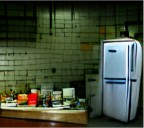}&
\includegraphics[width=0.08\linewidth,clip]
{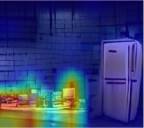}&
\includegraphics[width=0.08\linewidth,clip]
{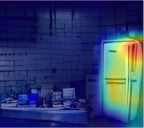}\\

\includegraphics[width=0.08\linewidth,clip]
{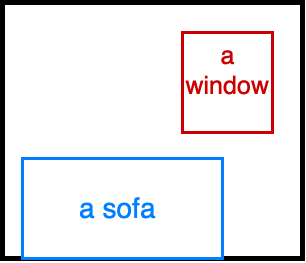}&
\includegraphics[width=0.08\linewidth,clip]
{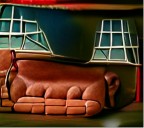}&
\includegraphics[width=0.08\linewidth,clip]
{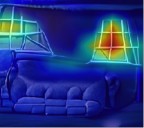}&
\includegraphics[width=0.08\linewidth,clip]
{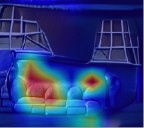}&
\includegraphics[width=0.08\linewidth,clip]
{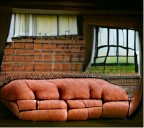}&
\includegraphics[width=0.08\linewidth,clip]
{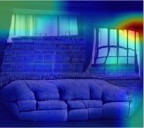}&
\includegraphics[width=0.08\linewidth,clip]
{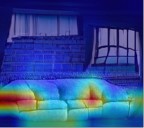}&
\includegraphics[width=0.08\linewidth,clip]
{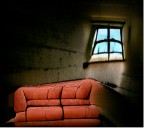}&
\includegraphics[width=0.08\linewidth,clip]
{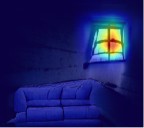}&
\includegraphics[width=0.08\linewidth,clip]
{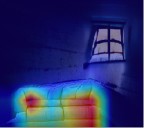}\\

\includegraphics[width=0.08\linewidth,clip]
{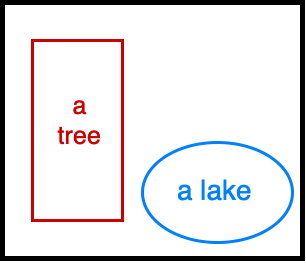}&
\includegraphics[width=0.08\linewidth,clip]
{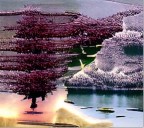}&
\includegraphics[width=0.08\linewidth,clip]
{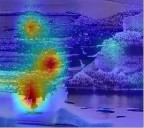}&
\includegraphics[width=0.08\linewidth,clip]
{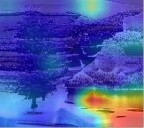}&
\includegraphics[width=0.08\linewidth,clip]
{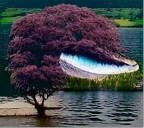}&
\includegraphics[width=0.08\linewidth,clip]
{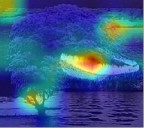}&
\includegraphics[width=0.08\linewidth,clip]
{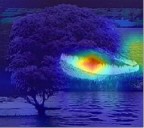}&
\includegraphics[width=0.08\linewidth,clip]
{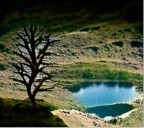}&
\includegraphics[width=0.08\linewidth,clip]
{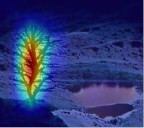}&
\includegraphics[width=0.08\linewidth,clip]
{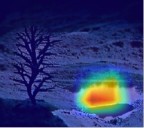}\\

\includegraphics[width=0.08\linewidth,clip]
{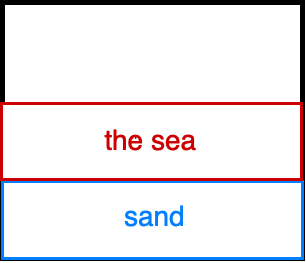}&
\includegraphics[width=0.08\linewidth,clip]
{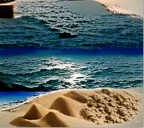}&
\includegraphics[width=0.08\linewidth,clip]
{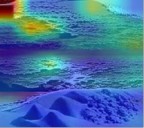}&
\includegraphics[width=0.08\linewidth,clip]
{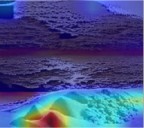}&
\includegraphics[width=0.08\linewidth,clip]
{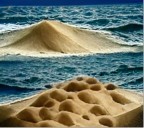}&
\includegraphics[width=0.08\linewidth,clip]
{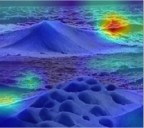}&
\includegraphics[width=0.08\linewidth,clip]
{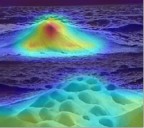}&
\includegraphics[width=0.08\linewidth,clip]
{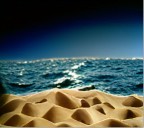}&
\includegraphics[width=0.08\linewidth,clip]
{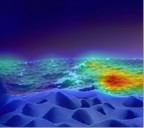}&
\includegraphics[width=0.08\linewidth,clip]
{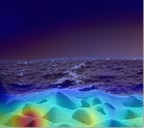}\\

\includegraphics[width=0.08\linewidth,clip]
{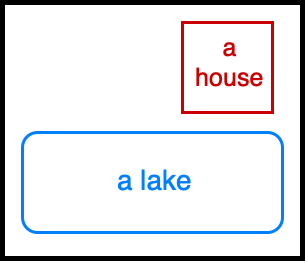}&
\includegraphics[width=0.08\linewidth,clip]
{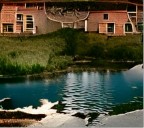}&
\includegraphics[width=0.08\linewidth,clip]
{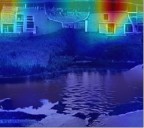}&
\includegraphics[width=0.08\linewidth,clip]
{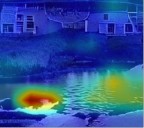}&
\includegraphics[width=0.08\linewidth,clip]
{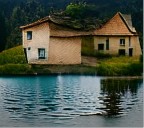}&
\includegraphics[width=0.08\linewidth,clip]
{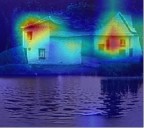}&
\includegraphics[width=0.08\linewidth,clip]
{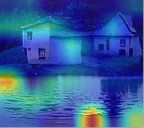}&
\includegraphics[width=0.08\linewidth,clip]
{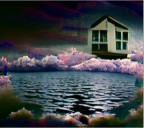}&
\includegraphics[width=0.08\linewidth,clip]
{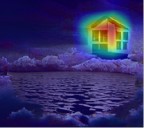}&
\includegraphics[width=0.08\linewidth,clip]
{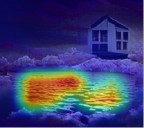}\\

\includegraphics[width=0.08\linewidth,clip]
{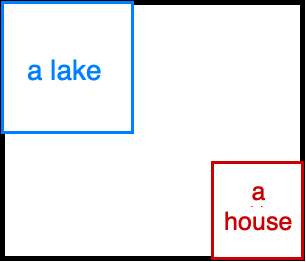}&
\includegraphics[width=0.08\linewidth,clip]
{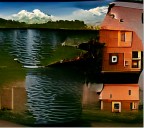}&
\includegraphics[width=0.08\linewidth,clip]
{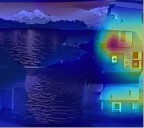}&
\includegraphics[width=0.08\linewidth,clip]
{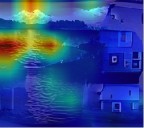}&
\includegraphics[width=0.08\linewidth,clip]
{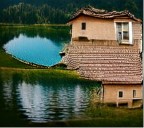}&
\includegraphics[width=0.08\linewidth,clip]
{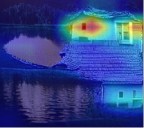}&
\includegraphics[width=0.08\linewidth,clip]
{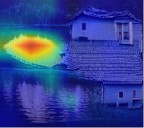}&
\includegraphics[width=0.08\linewidth,clip]
{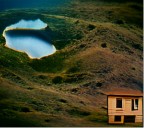}&
\includegraphics[width=0.08\linewidth,clip]
{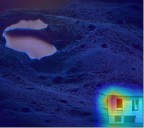}&
\includegraphics[width=0.08\linewidth,clip]
{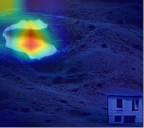}\\

\includegraphics[width=0.08\linewidth,clip]
{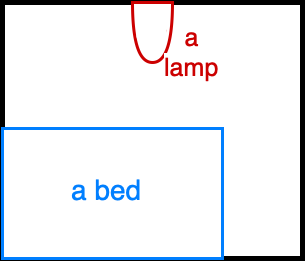}&
\includegraphics[width=0.08\linewidth,clip]
{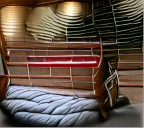}&
\includegraphics[width=0.08\linewidth,clip]
{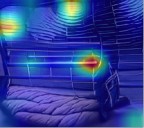}&
\includegraphics[width=0.08\linewidth,clip]
{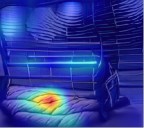}&
\includegraphics[width=0.08\linewidth,clip]
{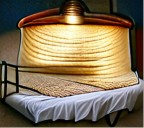}&
\includegraphics[width=0.08\linewidth,clip]
{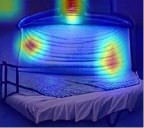}&
\includegraphics[width=0.08\linewidth,clip]
{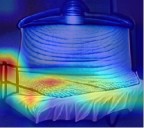}&
\includegraphics[width=0.08\linewidth,clip]
{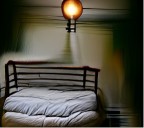}&
\includegraphics[width=0.08\linewidth,clip]
{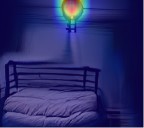}&
\includegraphics[width=0.08\linewidth,clip]
{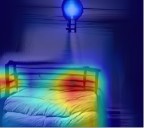}\\

\end{tabular}
\caption{Examples of spatially conditioned images generated with the similarity-based baselines and with our explainability-based method, along with the relevance maps produced for the images and their matching text prompts. 
The leftmost column presents the spatial conditioning provided to each method, each of the bounding boxes (red, blue) serves as a mask for the generation process.
For each method, the left image is the generated results, the middle image is the explainability map for the object conditioned with the red bounding box, and the right image is the explainability map for the object conditioned with the blue bounding box. As can be seen, the relevancy maps correspond well to the objects for our methods, while the baselines suffer from significant artifacts.\label{fig:l2i_with_expl}}
\end{figure*}

Fig.~\ref{fig:l2i_more_examples},~\ref{fig:l2i_more_examples2} present additional examples of our method for zero-shot text to image generation with spatial conditioning.  Fig.~\ref{fig:l2i_with_expl} presents the generated images along with their explainability maps.
Our explainability guidance enforces that the objects remain within the provided bounding boxes, and the relevancy maps demonstrate the effectiveness of the explainability method in detecting the objects within each box. As can be seen, {for most cases,} our method successfully generates the images such that each object is contained within its designated bounding box, while the similarity-based baselines tend to deviate from the provided bounding boxes, which is also reflected in artifacts in the explainability maps.

\begin{table}[t]
    \centering
   \begin{tabular*}{\linewidth}{@{\extracolsep{\fill}}lcccccc}
        \toprule
        $temp$ & Precision & Recall & F1 & AP & AR & AP$_{0.5}$\\
        \midrule
        1 & 52.1 & 72.8 & 53.6 & 8.4 & 21.3 & 23.4\\
        10 &  67.5 & \textbf{76.4} & \textbf{66.7} & 23.5 & 38.6 & 49.8\\
        20* & 71.7 & 63.4 & 62.6 & \textbf{26.2} & \textbf{40.0} & \textbf{56.5}\\
        30 & \textbf{72.6} & 53.0 & 56.8 & 23.3 & 36.0 & 52.2\\
        40 & 72.4 & 46.6 & 52.1 & 19.0 & 34.1 & 44.8\\
        \bottomrule
    \end{tabular*}
        \caption{{Precision, recall, F1, average precision, and average recall for spatially conditioned image generation with our method, with different values for the hyperparameter $temp$. Metrics are averaged across $100$ random samples from the MSCOCO~\cite{ty2014coco} validation set and four random seeds. Average precision and average recall are calculated using DETR~\cite{carion2020end}. * The value used for our method}}
    \label{tab:l2i_ablation_temp}
\end{table}

\begin{table}[t]
    \centering
   \begin{tabular*}{\linewidth}{@{\extracolsep{\fill}}lcccccc}
        \toprule
        $T$ & Precision & Recall & F1 & AP & AR & AP$_{0.5}$\\
        \midrule
        0.05 & 68.8 & \textbf{66.7} & 62.4 & 20.8 & 34.8 & 44.9\\
        0.1* & 71.7 & 63.4 & \textbf{62.6} & \textbf{26.2} & \textbf{40.0} & \textbf{56.5}\\
        0.2 & 73.4 & 56.3 & 58.6 & 22.2 & 37.2 & 52.2 \\
        0.3 & \textbf{73.7} & 53.8 & 56.7 &20.8 & 34.8 & 44.9\\
        0.5 & 71.5 & 58.1 & 57.3 & 19.3 & 33.4 & 41.4\\
        \bottomrule
    \end{tabular*}
        \caption{{Precision, recall, F1, average precision, and average recall for spatially conditioned image generation with our method, with different values for the threshold $T$. Metrics are averaged across $100$ random samples from the MSCOCO~\cite{ty2014coco} validation set and four random seeds. Average precision and average recall are calculated using DETR~\cite{carion2020end}. * The value used for our method}}
    \label{tab:l2i_ablation_T}
\end{table}

\begin{table}[t]
    \centering
   \begin{tabular*}{\linewidth}{@{\extracolsep{\fill}}lcccccc}
        \toprule
        $\lambda_{expl}$ & Precision & Recall & F1 & AP & AR & AP$_{0.5}$\\
        \midrule
        $\frac{0.1}{\sqrt{r(m)}}$ & 70.5 & 62,4 & 61.3 & 22.1 & 37.0 & 48.2\\
        $\frac{0.15}{\sqrt{r(m)}}$* & \textbf{71.7} & 63.4 & 62.6 & \textbf{26.2} & \textbf{40.0} & \textbf{56.5}\\
        $\frac{0.2}{\sqrt{r(m)}}$ & \textbf{71.7} & \textbf{64.1} & \textbf{62.7} & 24.4 & 38.5 & 51.3\\
        \bottomrule
    \end{tabular*}
        \caption{{Precision, recall, F1, average precision, and average recall for spatially conditioned image generation with our method, with different values for $\lambda_{expl}$. $r(m)$ denotes the ratio between the area of the mask and the area of the entire image. Metrics are averaged across $100$ random samples from the MSCOCO~\cite{ty2014coco} validation set and four random seeds. Average precision and average recall are calculated using DETR~\cite{carion2020end}. * The value used for our method}}
    \label{tab:l2i_ablation_lambda}
\end{table}

\end{document}